\documentclass{article}

\PassOptionsToPackage{numbers, compress}{natbib}
\usepackage[final]{neurips_2024}

\usepackage[utf8]{inputenc} % allow utf-8 input
\usepackage[T1]{fontenc}    % use 8-bit T1 fonts
\usepackage[colorlinks,
            linkcolor=red,
            anchorcolor=red,
            citecolor=blue
            ]{hyperref}      % hyperlinks
\usepackage{url}            % simple URL typesetting
\usepackage{booktabs}       % professional-quality tables
\usepackage{amsfonts}       % blackboard math symbols
\usepackage{nicefrac}       % compact symbols for 1/2, etc.
\usepackage{microtype}      % microtypography
\usepackage{xcolor}         % colors
\usepackage{graphicx} 

\usepackage{wrapfig}
\usepackage{nicefrac}       % compact symbols for 1/2, etc.
\usepackage{microtype}      % microtypography
\usepackage{xcolor}         % colors
\usepackage{mdframed}
\usepackage{pifont}
\usepackage{color}
\usepackage{wrapfig}
\usepackage{subfig}
\usepackage{makecell}
\usepackage{colortbl}
\usepackage{multirow}
\usepackage{fixltx2e}
\usepackage{subcaption}
\usepackage{etoolbox}
\usepackage{multicol}
\usepackage{refcount}
\usepackage{enumitem}
\usepackage[misc]{ifsym}
\usepackage{algorithm}
\usepackage{amsmath}
\definecolor{LightCyan}{rgb}{0.88,1,1}
\definecolor{mygray}{gray}{0.9}
\definecolor{mygray2}{gray}{0.6}
\mdfdefinestyle{boxstyle}{
    linewidth=1pt, % width of the frame
    linecolor=black, % color of the frame
    innerleftmargin=20pt, % space between frame and content on the left
    innerrightmargin=20pt, % space between frame and content on the right
    innertopmargin=20pt, % space between frame and content at the top
    innerbottommargin=20pt, % space between frame and content at the bottom
     backgroundcolor=gray!20
}

\title{Instruction-Guided Visual Masking}

\author{Jinliang Zheng\footnotemark[1]~~$^{1,2}$, Jianxiong Li\footnotemark[1]~~$^{1}$, Sijie Cheng$^{1}$, \textbf{Yinan Zheng}$^{1}$,\\ \textbf{Jiaming Li}$^{1}$, \textbf{Jihao Liu}$^{3, 2}$, \textbf{Yu Liu}$^{2}$, \textbf{Jingjing Liu}\footnotemark[2]~~$^{1}$, \textbf{Xianyuan Zhan}\footnotemark[2]~~$^{1, 4}$\\
$^1$ AIR, Tsinghua University, 
$^2$ Sensetime Research\\
$^3$ MMLab, CUHK,
$^4$ Shanghai AI Lab\\
\texttt{\{zhengjl23, li-jx21\}@mails.tsinghua.edu.cn}\\
\texttt{zhanxianyuan@air.tsinghua.edu.cn}\\
}

\begin{document}

\maketitle
\renewcommand{\thefootnote}{\fnsymbol{footnote}}
\footnotetext[1]{Equal contribution}
\footnotetext[2]{Corresponding author}

\vspace{-0pt}
\begin{abstract}
\vspace{-0pt}
Instruction following is crucial in contemporary LLM. However, when extended to multimodal setting, it often suffers from misalignment between specific textual instruction and targeted local region of an image. 
% Such visual grounding abilities in current LMMs grow implicitly through data-intensive end-to-end training on downstream tasks, with yet subpar performance. %yet even the most advanced LMMs trained on billions of data still exhibit subpar performance.
%This paper proposes a more direct and effective way to improve grounding in LMMs non-intrusively, by decoupling the challenging visual grounding problem thereby relieving existing models from this complex responsibility. \jj{This sentence doesn't mean anything}
% , that is first ask visual grounding models to focus on task-relevant visuals and then do multimodal instruction following based on the highlighted visual contents. 
To achieve more accurate and nuanced multimodal instruction following, we introduce \textit{Instruction-guided Visual Masking} (IVM), a new versatile  visual grounding model that is compatible with diverse multimodal models, such as LMM and robot model.
% specifically designed for LMM.  
%\zjl{"we introduce Instruction-guided Visual Masking(IVM), a new visual grounding module specifically designed for LMM" may be more accurate.}
%\jj{Is IVM a new task or a new method?}
%\ljx{We think we would better not call it a new task because we may have to benchmark this new task. IVM can be seen as an extension of existing visual grounding method to a more complex setting where the instruction is open-world vacabulary.}
%to cover the broader and complex multimodal domain for instruction following. 
By constructing visual masks for instruction-irrelevant regions, IVM-enhanced multimodal models can effectively focus on task-relevant image regions to better align with complex instructions. %However, due to the 
Specifically, we design a visual masking data generation pipeline and create an IVM-Mix-1M dataset with 1 million image-instruction pairs. We further introduce a new learning technique, \textit{Discriminator Weighted Supervised Learning} (DWSL) for preferential IVM training that prioritizes high-quality data samples.
% \zh{Maybe change IL to supervised learning?},  %\zjl{this sentence may raise some misunderstanding as IVM is an extra module decoupled from LMMs, and DWIL is introduced to boost the training efficiency of IVM} \jj{Better?} \zjl{yeah, but " new visual grounding method specifically designed for LMM training." may be confused either.} 
Experimental results on generic multimodal tasks such as VQA and embodied robotic control demonstrate the versatility of 
% the proposed IVM model,
IVM,
which as a plug-and-play tool, significantly boosts the performance of diverse multimodal models, yielding new state-of-the-art results across challenging multimodal benchmarks. Code, model and data are available at \href{https://github.com/2toinf/IVM}{\texttt{https://github.com/2toinf/IVM}}.

\end{abstract}

\vspace{-0pt}
\section{Introduction}
\begin{figure}[t]
    \centering
    % \vspace{-8pt}
    \includegraphics[width=0.99\textwidth]{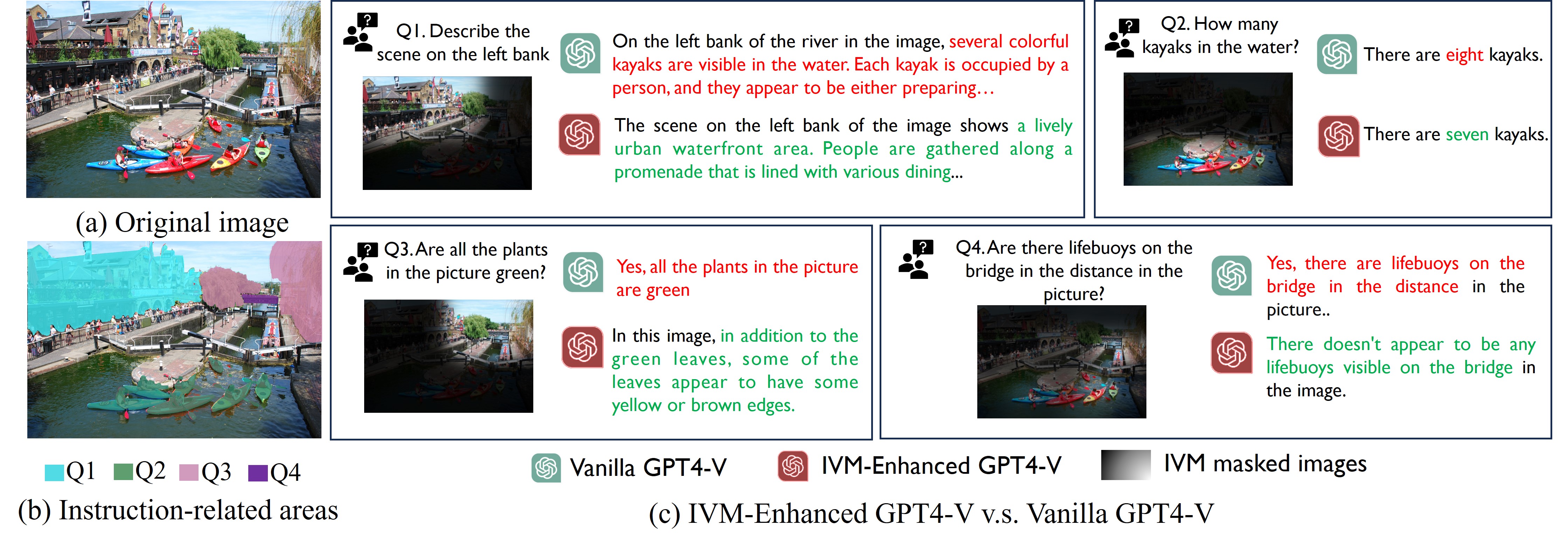}
    % \vspace{-5pt}
    \caption{\small The most advanced LMMs (e.g. GPT4-V) still fail on complex instruction following tasks. With IVM assistance to simplify visual inputs, existing LMMs can gain significant improvement.}
    \label{fig:lmm_fail}
    % \vspace{-3pt}
    \centering
    % \hspace{pt}
    \includegraphics[width=0.99\textwidth]{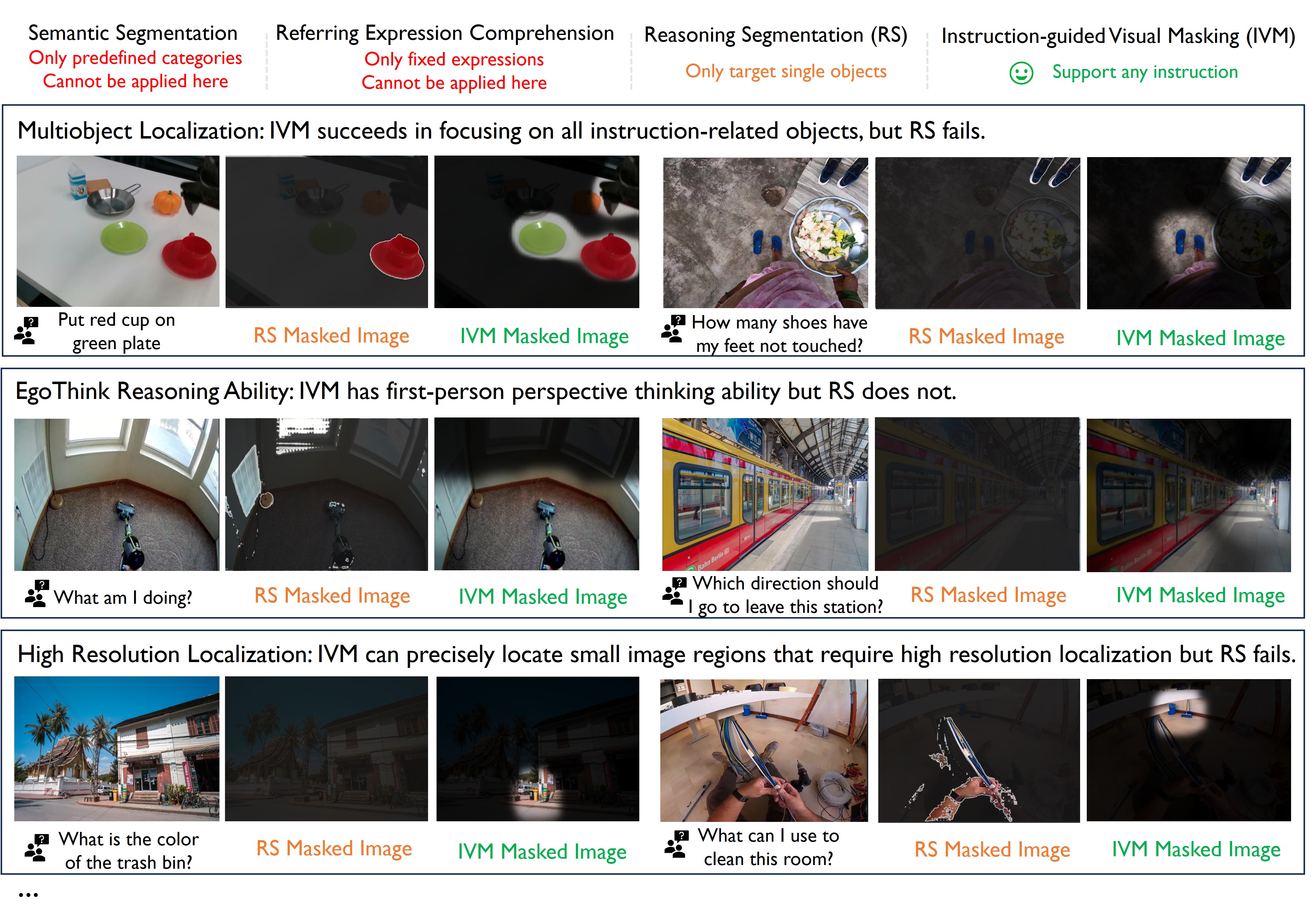}
    % \vspace{-5pt}
    \caption{\small 
    Comparison between IVM and Reasoning Segmentation (RS)~\cite{lai2024lisa}. Traditional methods such as semantic segmentation~\cite{ade20k} and referring expression comprehension~\cite{yu2016rec} are limited to fixed categories or fixed instruction formats, thus inapplicable to complex instruction following tasks.
    % \jj{If only RS is compared, this description shouldn't be here. It's weird to include the frown/happy faces here. Perhaps put it into a table} .
    RS has reasoning ability, but only allows single object localization. IVM, instead, is universally applicable to any instruction.%
    % \jj{Still typo: succeed -> succeeds in focusing. Why are there half brackets around the 4 methods?}\ljx{because IVM can cover RS, RS can cover referring expression comprehension, which further covers semantic segmentation} \jj{There's no way people can relate to this by looking at the half brackets} 
    %\ljx{Got it, maybe I can delete these brackets to avoid confusion here. These relations are not the core part of this figure.}
    % \jj{Perhaps add dotted lines as separators between them}\ljx{OK!}
    }
    % \jj{Text in the figure has typos. e.g., 'success' should be 'succeed'. 'RS not' should be 'RS does not'. 'High resolution localization' should be capitalized like the other two tasks}}
   
    % \zh{enlarge font size, words are not very readable}\ljx{OK}}
    % \vspace{-10pt}
    \label{fig:intro}
\end{figure}

\vspace{-0pt}
Multimodal instruction following is a fundamental multimodal task,
%Large Multimodal Models (LMM) serve as a generalist 
powering a wide-range of applications such as visual question answering (VQA)~\cite{goyal2017vqav2}, visual captioning~\cite{gpt4, llava}, and embodied robotic control~\cite{driess2023palm}. 
% To train an effective and scalable LMM, 
To effectively solve this task,
one critical capability required is nuanced image-language grounding, which current multimodal models grow implicitly and slowly through data-intensive end-to-end training without explicit grounding supervisions. Two challenges emerge in this indirect learning of image-instruction alignment: $1)$ How to accurately localize targeted image regions that corresponds to a specific textual instruction, as illustrated in Figure~\ref{fig:lmm_fail}. $2)$ How to generalize to diverse visual representations (\textit{e.g.}, same object with different colors, compositions, or backgrounds) that reflect similar textual instruction (e.g., Q3 in Figure~\ref{fig:lmm_fail}).  % instruction-related visual contents may vary significantly w.r.t different instructions, and extraneous visuals can introduce substantial distractions.  
% are most of the current advanced LMM are end-to-end trained on task-specific data to enable the corresponding generation while emerging visual grounding and understanding capability implicitly.
% The end-to-end training paradigm, however, 
Lacking an effective and direct solution to these challenges, the most advanced Large Multimodal Models (LMMs)~\cite{gpt4, bai2023qwen, llava, driess2023palm}
% as the most powerful LMMs~\cite{gpt4} trained with massive data 
still suffer from hallucinations even when trained with high-quality data in the magnitude of billions~\cite{li2023pope}.
% , especially given complex visual inputs. 
%Therefore, in this paper, we aim to explore a more data-efficient and direct approach to inject strong visual grounding ability to existing LMM non-intrusively.

We introduce \textit{Instruction-guided Visual Masking} (IVM), a versatile plug-and-play model designed to enhance 
% the performance of advanced LMMs 
multimodal instruction following 
% Unlike existing visual grounding methods that are limited to narrow instructions or simple visual targets~\cite{lai2024lisa} (Figure~\ref{fig:intro}), IVM is suitable for broader and more complex instruction-following tasks.
via nuanced surgical visual grounding. To eliminate the distraction of instruction-irrelevant visual regions, IVM automatically masks out these regions to sharpen the focus of instruction following, and meticulously crops visual input to tailor for a specific instruction and enforce multimodal models to zoom in on task-related visual content. %This approach decouples the challenging visual grounding problems from LMMs, allowing them to follow fine-grained and complex instructions without distraction.
Existing visual grounding methods are limited either to predefined object categories, which cannot cover diverse instruction-related visual content; or they subscribe to a fixed instruction format, which restricts the expressiveness of instructions. As shown in Figure~\ref{fig:intro}, such simplistic grounding techniques often fail to comprehend complex instruction-following tasks.
% \jj{The comparison with other visual grounding tasks is inaccurate. What does it mean 'narrow' and 'simple'? Need better explanation on why existing visual grounding tasks are not sufficient for LMM training.}

%Although some attempts have studied similar problems~\cite{lai2024lisa}, this area remains under-explored, with central challenges stemming from limited data and suboptimal training algorithms. No existing datasets are specifically designed for IVM training, as it 
Learning an IVM model requires pixel-level, fine-grained, instruction-guided mask annotations that provide explicit grounding supervisions. To create such a dataset, we build a LLM-empowered Mixture of Expert pipeline with SOTA visual grounding models~\cite{sun2023alphaclip, shao2024visualcot, lai2024lisa, owdetr} to efficiently create abundant reliable labels. %However, due to IVM task surpasses the capabilities of all existing LMM, auto-generated labels contain noises that fail to completely exclude irrelevant visuals or mistakenly filter out critical contents. Thus, 
To compensate the noises in auto-generated labels, we further manually label a smaller dataset with clean annotations, and integrate the two into an IVM-Mix-1M dataset that contains 1 million image-instruction pairs. %This dataset provides explicit grounding supervision, which can be readily absorbed into IVM model to enhance LMM performance. We show that the IVM model trained directly on IMV-Mix-1M can introduce obvious improvement for LMM.

%However, training solely on the mixed-quality data is inferior since some inaccuracies exist in auto-generated labels. To remedy this and fully unleash the power of the IVM-Mix-1M dataset, we adopt a novel  Discriminator-Weighted Imitation Learning (DWIL) framework for IVM training, 
To reduce demand on costly human labels and ensure optimized utility of machine-generated labels, we employ a Discriminator-Weighted Supervised Learning (DWSL) framework for IVM training, inspired by recent advances in offline imitation learning~\cite{xu2022discriminator}. Specifically, we introduce a discriminator
% trained by Positive-Unlabeled (PU) learning~\cite{du2015convex} 
to assign weights to masks, where high values are assigned to high-quality annotations and vice versa. Thus, these weights generated by the discriminator can naturally act as a weighting function for the IVM training objective, allowing for a preferential training process that prioritizes learning from reliable samples and discards misleading ones. %This approach greatly reduces the demands for costly human labels, as well as addresses the issues associated with machine-generated labels.

Extensive experiments demonstrate great versatility of the IVM model when integrated into existing multimodal chatbots (commercial and open-sourced) without fine-tuning.
% This integration significantly enhances the performance of these models and
Our IVM-enhanced LMMs gain significant performance improvement across new challenging benchmarks such as  V*Bench~\cite{wu2023vstar}, EgoThink~\cite{cheng2024egothink} and POPE~\cite{li2023pope}, achieving new state of the art. IVM model also proves valuable in vision-language robotic manipulation tasks, where data collection is notoriously challenging and generalization is a major concern~\cite{li2024decisionnce}. With the integration of IVM, our enhanced robot model exhibits boosted performance and better generalization capabilities.

Our contributions are summarized as follows:
$1)$ We propose Instruction-guided Visual Masking (IVM), a novel approach that serves as a versatile plug-and-play module to enhance multimodal models through visual grounding.
$2)$ We introduce the IVM-Mix-1M dataset and propose an LLM-empowered Mixture of Expert pipeline to create visual grounding labels. % We also manually label a subset to further enhance the label quality.
$3)$ We present the DWSL algorithm for IVM training that automatically prioritizes high-quality training samples. %Extensive experiment results demonstrate the model's effectiveness.

\vspace{-0pt}
\section{Related Work}
\vspace{-0pt}
\textbf{Large Multimodal Models}. \ LLaVA~\cite{llava} first demonstrates promising capabilities in following complex instructions. %Large MultiModal Models (LMMs) have rapidly gained significant attention across a wide range of fields.
Subsequent works such as LLaVA-1.5~\cite{llava_1_5}, MiniGPT4~\cite{zhu2023minigpt} Qwen-VL~\cite{bai2023qwen} and CogVLM~\cite{wang2023cogvlm}, further enhance LMMs via refined model design and enriching the quality of training data, achieving state-of-the-art performance on diverse downstream tasks including visual grounding~\cite{liang2019vrrvg}, visual reasoning~\cite{thrush2022winoground}, visual question and answering~\cite{goyal2017vqav2}. Moreover, by integrating the robotics action modality, LMMs perform versatile planning and manipulation in instruction-driven robotics tasks. Notable studies in this line of inquiry include PaLM-E~\cite{driess2023palm}, the series of RT models~\cite{brohan2022rt, brohan2023rt, rtx}, and text-guided video planning diffusion models~\cite{du2023learning, yang2024learning, black2024zeroshot}. Despite the success, LMMs still struggle with complex visual grounding challenges, often misreading instruction-irrelevant visual contents (Figure~\ref{fig:lmm_fail}). To address this, researchers have tried to adapt existing visual modules to higher-resolution images to obtain better perception~\cite{liu2024llavanext}, but with limited improvement.

\textbf{Visual Grounding Tasks}.
Visual grounding requires precisely localizing image regions corresponding to a referring expression, among which the RefCOCO series~\cite{yu2016rec} is the most well-known benchmark, 
% The community has witnessed a surge in the amount of
and numerous public visual grounding data are available~\cite{liang2019vrrvg, young2014flickr, gupta2019lvis}.
% , such as Visual Genome~\cite{liang2019vrrvg}, Flicker30K~\cite{young2014flickr} and LVIS~\cite{gupta2019lvis}.
Recently, LMMs incorporate these visual grounding data via visual-instruction tuning~\cite{llava, llava_1_5, zhu2023minigpt}, establishing new SOTA in this area~\cite{shao2024visualcot}.
% Recently, these data are integrated as a component of visual instruction tuning~\cite{llava, llava_1_5, zhu2023minigpt} within LMMs, which swiftly establish themselves as state-of-the-art~\cite{shao2024visualcot} in this area. 
To further broaden the reasoning ability of visual grounding,  LISA~\cite{lai2024lisa} introduces a new task, reasoning segmentation, which demands higher capabilities in instruction comprehension. However, visual grounding is still limited to align simple instruction with specific objects, which cannot adapt to more complex instruction following tasks (\textit{e.g.} Figure~\ref{fig:intro}).

% \subsection{Visual Grounding}
\textbf{Visual Grounding Augmented LMMs}. \ Recently, a series of visual grounding methods emerged to enhance the performance of LMMs in complex visual scenes. V*~\cite{wu2023vstar} employs a heuristic search strategy to search, locate, and crop image areas relevant to instructions through a multi-step iterative process. VisualCot~\cite{shao2024visualcot} is trained end-to-end with a customized dataset to achieve target localization capabilities. These two methods allow LMMs to dynamically focus on visual inputs until the correct answer is derived. However, these complex inference pipelines lead to substantial computational overhead, and their heuristic designs further hinder the extension beyond VQA to other multimodal instruction following tasks such as robotic control. 

Besides these explicit strategies incorporating additional visual grounding modules, other studies pursue refining data or introducing extra training targets to enhance the grounding capabilities of LMMs implicitly. ViGor~\cite{yan2024vigor} proposes a fine-grained reward modeling to enhance visual grounding of LMMs, and 
% , but the grounding ability is still indirectly incorporated via a learned reward model on downstream tasks, which can be highly data-inefficient~\cite{hu2024querypolicy}.
SynGround~\cite{he2024learning} introduces a pragmatic framework for image-text-box synthesis tailored for visual grounding. These methods, however, are primarily focused on the visual grounding task itself, overlooking its influence on downstream multimodal instruction following tasks.
% can only provide coarse-grained rectangular grounding results, failing to achieve the pixel-level and fine-grained grounding that IVM offers in this paper. 

Distinct from previous efforts, this paper introduces a generic visual grounding model that is adaptable to any multimodal instruction following tasks, and provides a systematic investigation into the advantages of integrating an additional visual grounding model into downstream applications.
% directly focusing on pixel-level regions of interest as dictated by the instructions in an effective and straightforward way.
% Despite the impressive performance of these innovations, they share significant limitations: the 
% introduction of task-specific data, 
% complex inference pipelines and substantial computational overhead greatly limits their application scope.

\section{Instruction-Guided Visual Masking}

To help multimodal models focus on instruction-sensitive image regions without distractions from irrelevant visual elements, we introduce Instruction-guided Visual Masking (IVM), a versatile plug-and-play model that enhances multimodal instruction following via surgical targeted visual grounding.

% \vspace{-4pt}
\subsection{Problem Definition}
% \vspace{-4pt}
\label{subsec:problem_def}

\begin{wrapfigure}{r}{0.47\textwidth} 
\vspace{-10pt}
\centering
\includegraphics[width=0.45\textwidth]{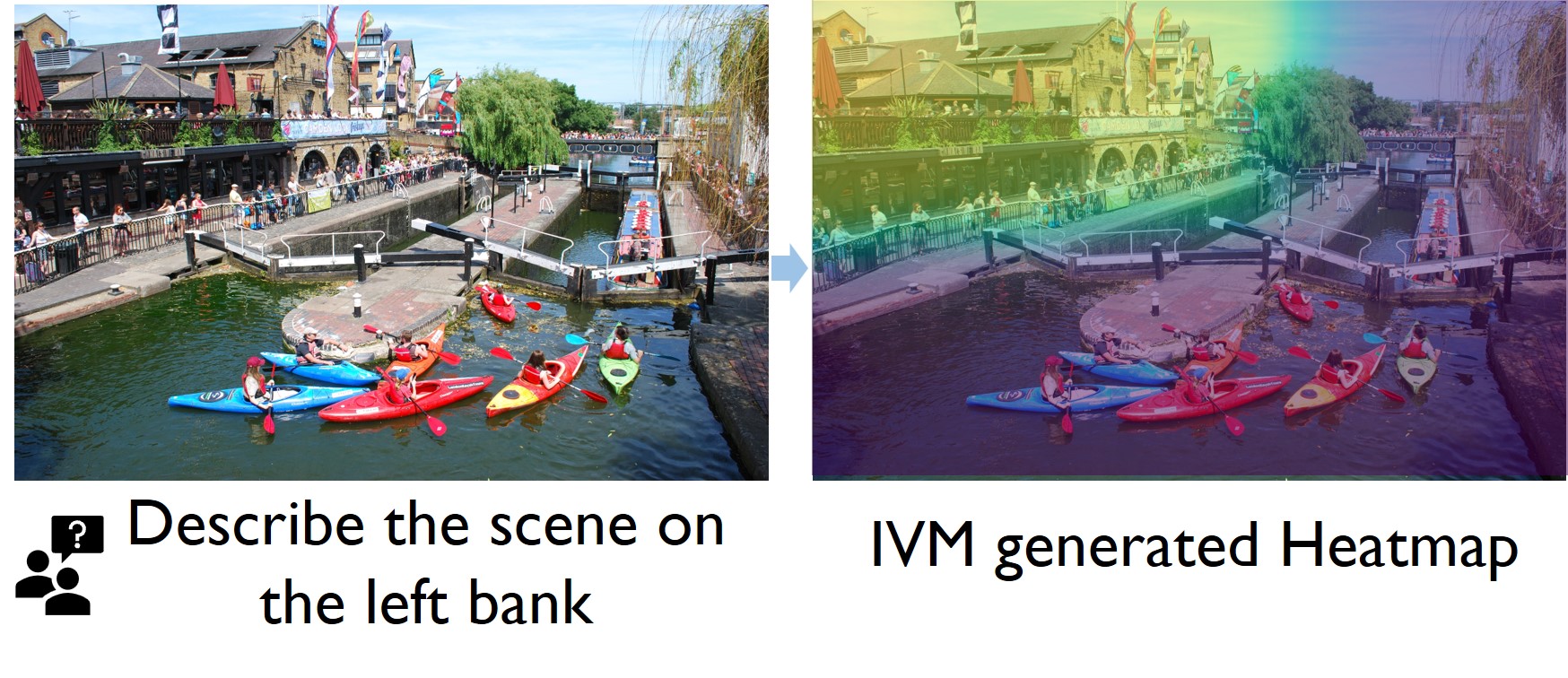}
\vspace{-5pt}
\caption{\small Instruction-guided Visual Masking.}
\label{fig:ivm_def}
\vspace{-10pt}
\end{wrapfigure}
IVM aims to produce a heatmap $\mathbf{H}$, given an image $\mathbf{x}_{\rm img}$ and a textual instruction $\mathbf{x}_{\rm txt}$. The heatmap $\mathbf{H}$ identifies the critical image region to follow the instructions, as illustrated in Figure~\ref{fig:ivm_def}, allowing multimodal models to easily zoom in on targeted image regions while ignoring neighboring areas.

This formulation evokes the problem definition of Reasoning Segmentation (RS)~\cite{lai2024lisa}. There are two main differences: $1)$ IVM addresses a more challenging problem. RS tries to target single objects from simple instructions, \textit{e.g.}, "what is.., where is.., who is...", while IVM aims to include all instruction-related visual regions within the image given any instruction, which demands advanced and nuanced image-language grounding ability (as illustrated in Figure~\ref{fig:intro}). $2)$ RS has clear ground truths but IVM does not. The instructions in RS primarily correspond to simple and semantic-meaningful objects that are straightforward for human annotations. IVM, however, deals with broader and more ambiguous instruction-related regions (\textit{e.g.}, the left bank regions in Figure~\ref{fig:ivm_def}), making the training and annotating much more challenging.

% \vspace{-3pt}
\subsection{Data Preparation}
% \vspace{-3pt}
\begin{figure*}[t]
    \centering
    \includegraphics[width=0.99\textwidth]{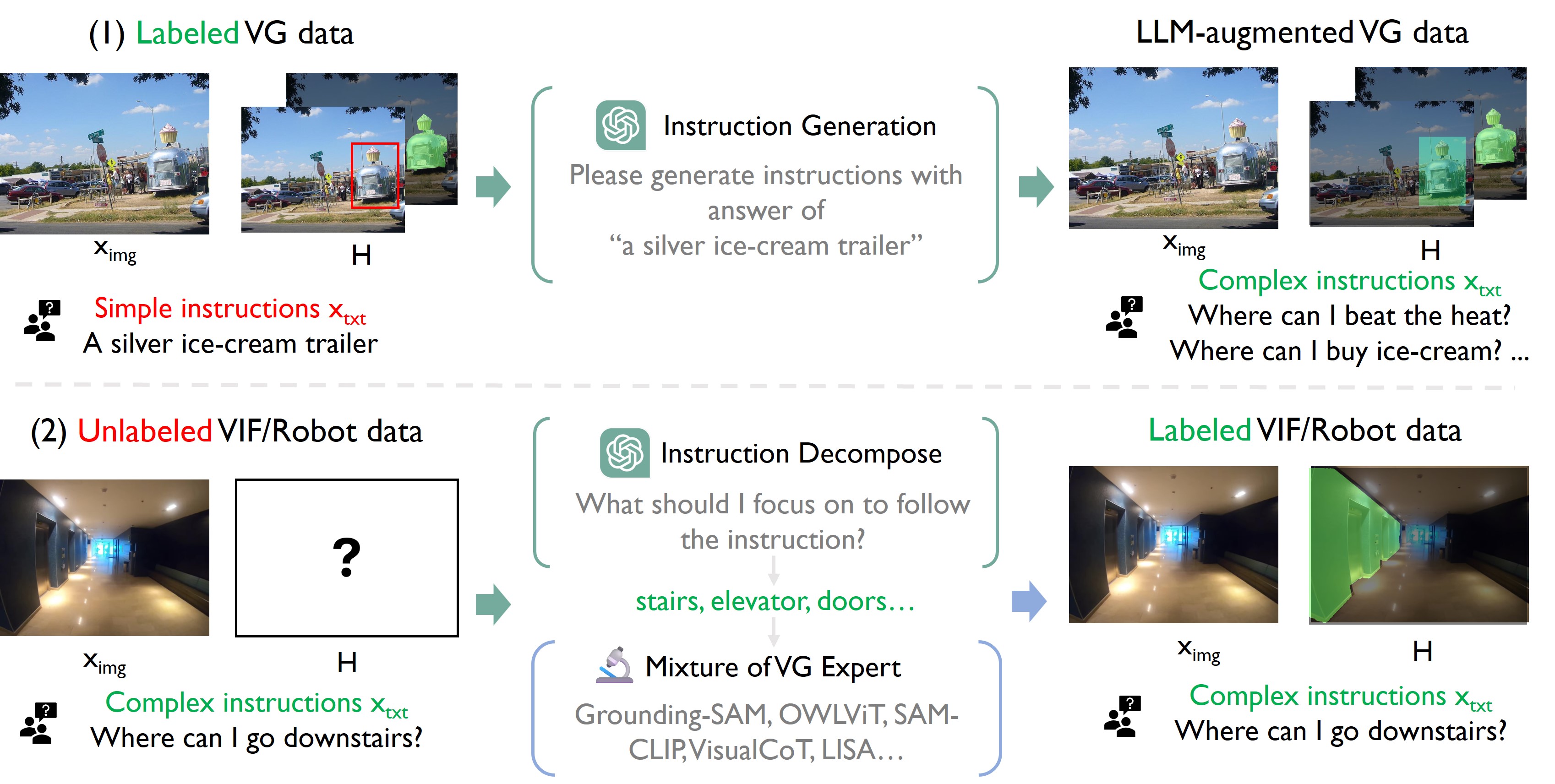}
    % \vspace{-5pt}
    \caption{\small LLM-empowered Mixture-of-Expert pipeline for auto-annotation. 
    (1)~For labeled VG data, we utilize an LLM to generate complex instruction annotations.
    (2)~For unlabeled VIF or robot data, we first use an LLM to simplify the instruction and then leverage a mixture of VG models to generate candidate labels.
    %\jj{Add some explanation for the figure, e.g., differences between the images}
    }
    % \vspace{-5pt}
    \label{fig:data_prepare}
\end{figure*}

% LLM-empowered Mixture of Expert data generation pipeline}
To train an IVM model, the first main challenge is the scarcity of training data. Most existing Visual Grounding (VG) datasets~\cite{yu2016rec, lai2024lisa} typically feature simple instructions focused primarily on prominent objects within images, lacking both diversity and complexity required for IVM. To tackle this, we compiled one million data from various sources, including labeled visual grounding, unlabeled multimodal instruction following, and robotics data. As outlined in Section~\ref{subsec:problem_def}, scaling human annotations is challenging due to the high complexity of such data. Therefore, we introduce an \textit{LLM-empowered Mixture of Expert
pipeline} that integrates SOTA visual grounding models to efficiently generate reliable annotations. We further manually annotate a smaller dataset to compensate inaccuracies in auto-generated labels. The resulted combined dataset, IVM-Mix-1M, comprises one million data samples ready for IVM training, which can be found in \href{https://github.com/2toinf/IVM}{\texttt{https://github.com/2toinf/IVM}}.
% By leveraging large language models and integrating existing visual grounding models, we generate high-quality grounding and language annotations. Given the complexity of some instructions which may lead to inaccuracies, we manually annotate a subset to further refine the dataset's quality. Detailed methods used for data construction for each category will be discussed in subsequent sections.

\textbf{LLM-empowered Mixture of Expert Annotation Pipeline.} Leveraging the power of LLM, this pipeline can efficiently generate high-quality annotation, which consists of two components (Figure~\ref{fig:data_prepare}): $1)$ ~\textit{Labeled visual grounding data.} We collect 250K labeled \textit{VG} data from multiple sources including VG caption~\cite{liang2019vrrvg}, Flickr30K~\cite{young2014flickr}, VSR~\cite{alayrac2019visual}, OpenImage~\cite{openimage}, and RefCoCo~\cite{yu2016rec, coco}, which provide bounding boxes with simple instructions for each image. To increase the diversity and complexity of instructions, 
%To achieve more fine-grained labeling, we use the \textit{Segment Anything Model (SAM)} to generate segmentation maps for the main objects within these bounding boxes. Following~\cite{llava}, 
we utilize GPT-4~\cite{gpt4}, known for its robust language understanding and generation capabilities, to create diverse instruction-answer pairs based on  existing language instructions. $2)$~\textit{Unlabeled {Visual-Instruction-Following (VIF)} and \textit{robotics} data.} We sample a 700K subset from LLaVA-Instruction-tuning~\cite{llava} for VQA-type data, and a 50K subset from OpenX~\cite{rtx} for robotics data. Given that these data lack grounded labels but contain complex instructions, we use GPT-4 to simplify the language instructions by prompting it to infer the names of targeted objects necessary for following the instructions. These simplified instructions then guide existing VG models to generate candidate labels. To ensure the quality of these labels and compensate for the ambiguous nature of the IVM task, we integrate proposals from several \textit{VG} experts, such as Grounding-Sam~\cite{ren2024grounded}, LISA~\cite{lai2024lisa}, AlphaClip~\cite{sun2023alphaclip}, and OwlViT~\cite{owdetr}, via an ensemble approach.

\textbf{Manual Annotation}. Despite integrating the most advanced models, the auto-generation design still faces challenges that can lead to data inaccuracies. First, employing LLM to simplify or complicate language annotations without considering image content can introduce uncontrollable biases. Second, as the task exceeds the capabilities of existing models, it becomes impossible to totally exclude low-quality annotations that contain irrelevant visuals or mistakenly filter out critical contents. Thus, to enhance the overall quality of the dataset, we further manually annotate a 10K subset of the constructed dataset to inject human expert knowledge. 

\begin{wrapfigure}{r}{0.47\textwidth} 
\vspace{-10pt}
\centering
\includegraphics[width=0.45\textwidth]{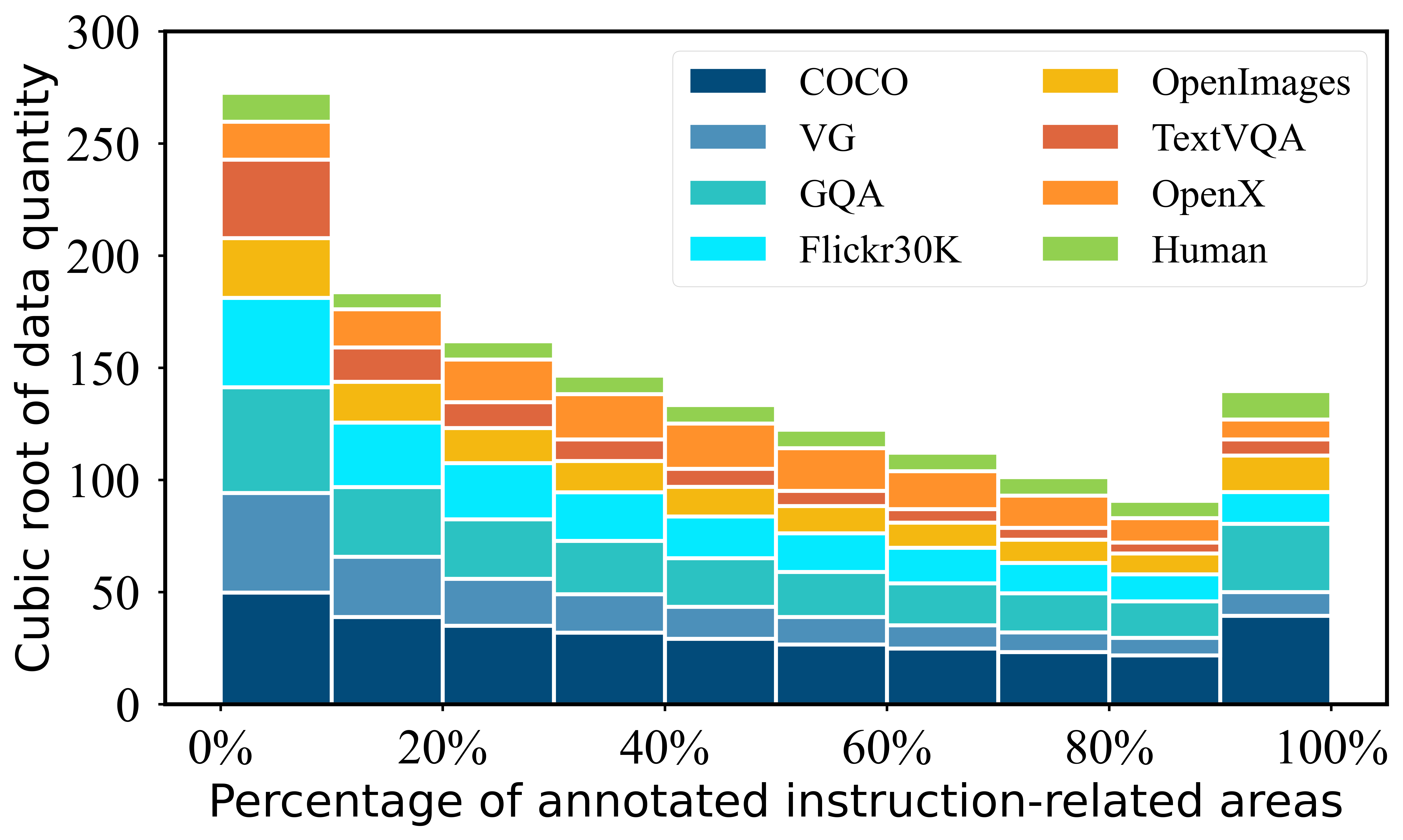}
\vspace{-5pt}
\caption{Data analysis on the IVM-Mix-1M dataset: data quantity v.s percentage of instruction-related areas.}
% \jj{Ratio should be percentage in the figure legend}}
\label{fig:data}
\vspace{-10pt}
\end{wrapfigure}
\textbf{Data Analysis}. 
% To better illustrate the composition of the IVM-Mix-1M dataset, 
Here, we provide quantitative analysis on the IVM-Mix-1M dataset. Figure~\ref{fig:data} depicts the data quantities w.r.t the percentage of annotated instruction-related image area. 
% We visualize 
Here, each ratio range is further categorized by different data sources, where manually annotations are treated as a separate category (Human), while all others are machine-generated. Our analysis reveals that the instruction-related image regions only occupy a small fraction of the total image area (\textit{e.g.} most data have less than 40\% instruction-relevant image regions), indicating that most visual contents may cause distraction  
% in a significant portion of the data, the area of the instruction-related image regions occupies a very small fraction of the total image area, whether annotated manually or by machine.
and corroborating the necessity of visual masking for instruction following tasks.

% \vspace{-7pt}
\subsection{Discriminator-Weighted Supervised Learning Framework}
% \vspace{-7pt}
The challenge now is to train the IVM model with a small high-quality human-annotated dataset ($\mathcal{D}_e$) as well as a large but mixed-quality auto-generated dataset ($\mathcal{D}_o$). Training naively on the combined dataset may yield suboptimal results due to inaccuracies in auto-generated labels, while solely using limited human-annotated data is insufficient. 
% provide satisfied performances alone. 
Inspired by recent advances in imitation learning using mixed-quality data~\cite{xu2022discriminator,zhang2023discriminator}, we 
% To reduce the highly demands on human annotations, as well as diminish the side-effects of machine-generated data, 
employ a Discriminator-Weighted Supervised Learning (DWSL) framework to effectively leverage the strengths of both auto- and human-annotated data.
% , balancing their respective strengths and weaknesses. 

\begin{figure}[t]
    \centering
    \includegraphics[width=0.99\textwidth]{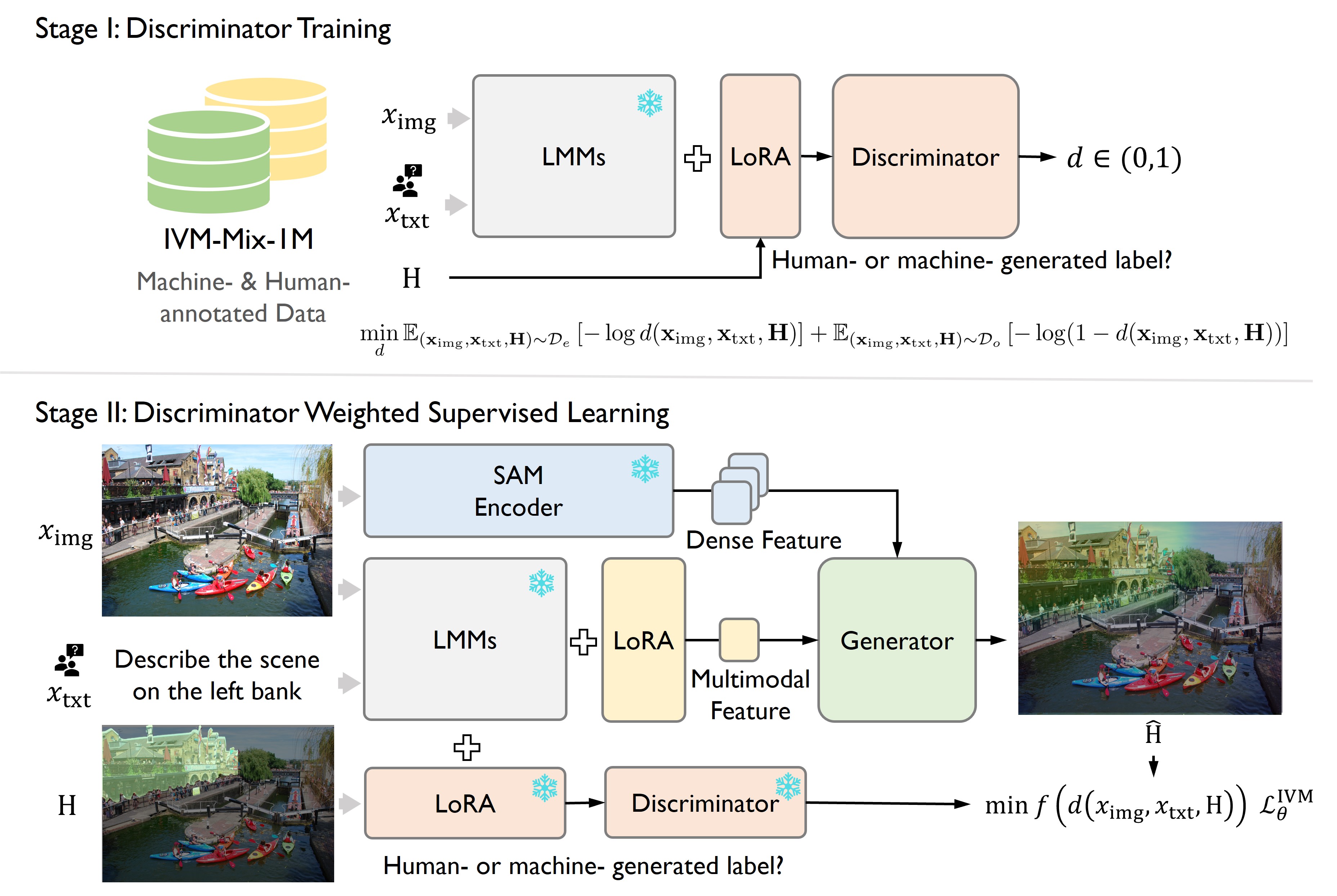}
    % \vspace{-3pt}
    \caption{\textbf{IVM model architecture and training pipeline.} Stage I: A LoRA-tuned LMMs is trained to discriminate human- and machine-annotated data. Stage II: A frozen SAM vision backbone and a LoRA-tuned LMMs are utilized to extract dense image features and multimodal representations, respectively. These features are then fed into a generator for dense prediction and is trained via DWSL. Same color represents the same model. See Appendix~\ref{subsec:arch_appenx} for more details.}
    % IVM model architecture. A frozen vision backbone and a LoRA-tuned LMM are utilized to extract dense image features and multimodal representations, respectively. These features are then fed into a generator for dense prediction and a discriminator for generating training weights. }
    
    %\jj{Should add some explanation for the figure, instead of saying 'see appendix'}
    \label{fig: architecture}
    % \vspace{-5pt}
\end{figure}

\textbf{Discriminator Training}. Specifically, we introduce a discriminator $d$ optimized by Eq.~(\ref{equ:discri})
% optimized by Positive-Unlabeled (PU) learning~\cite{du2015convex}
to assign high weights to high-quality annotations and vice versa:
% \textbf{Discriminator Training}. The problem now is how to achieve the best possible performance given a small but high-quality dataset ($\mathcal{D}_e$) and a large but mixed-quality dataset ($\mathcal{D}_o$), which is widely studied in offline IL~\cite{xu2022discriminator} and offline RL~\cite{levine2020offline}, but is highly under-explored in the settings of LMMs and multimodal training. 
% Specifically, we introduce a discriminator $d$ trained by Positive-Unlabeled (PU) learning~\cite{du2015convex} to assign weights based on annotation qualities:
\begin{equation}
\begin{aligned}
    \min_d \mathbb{E}_{(\mathbf{x}_{\rm img}, \mathbf{x}_{\rm txt}, \mathbf{H})\sim\mathcal{D}_e}&\left[-\log d(\mathbf{x}_{\rm img}, \mathbf{x}_{\rm txt}, \mathbf{H})\right] + \mathbb{E}_{(\mathbf{x}_{\rm img}, \mathbf{x}_{\rm txt}, \mathbf{H})\sim\mathcal{D}_o}\left[-\log (1-d(\mathbf{x}_{\rm img}, \mathbf{x}_{\rm txt}, \mathbf{H}))\right], \\
    % &-\alpha \mathbb{E}_{(\mathbf{x}_{\rm img}, \mathbf{x}_{\rm txt}, \mathbf{H})\sim\mathcal{D}_e}\left[-\log (1-d(\mathbf{x}_{\rm img}, \mathbf{x}_{\rm txt}, \mathbf{H}))\right],
\end{aligned}
\label{equ:discri}
\end{equation}
% where $(\mathbf{x}_{\rm img}, \mathbf{x}_{\rm txt}, \mathbf{H})$ are image-instruction-heatmap pairs sampled from $\mathcal{D}_o$ and $\mathcal{D}_e$ datasets. After training on Eq.~(\ref{equ:discri}), the discriminator $d$ can assign high weights for high-quality human-annotated data from $\mathcal{D}_e$ and relatively high weights for data hidden in $\mathcal{D}_o$ that behaves similar to $\mathcal{D}_e$ that conforms to human preferences.
where $(\mathbf{x}_{\rm img}, \mathbf{x}_{\rm txt}, \mathbf{H})$ are image-instruction-heatmap pairs sampled from $\mathcal{D}_o$ and $\mathcal{D}_e$ datasets. Eq.~(\ref{equ:discri}) is similar to the one in GAN~\cite{goodfellow2014generative}, but the "fake" data in ~\cite{goodfellow2014generative} is replaced by machine-generated data from $\mathcal{D}_o$.
After training with Eq.~(\ref{equ:discri}), the discriminator $d$  assigns high weights to high-quality human-annotated data from $\mathcal{D}_e$ and relatively high values to similarly high-quality data from $\mathcal{D}_o$ that aligns with human preferences, acting as a judge for annotation quality.
% $\hat{\mathbf{H}}$ are detached IVM generated heatmaps. The discriminator here can 
% Note the ground truth heatmap $\mathbf{H}$ cannot be obtained in practice due to the complexity as described in Section~\ref{subsec:problem_def}. So, $\mathbf{y}$ are binary masks that are crowdsourced annotated by humans and different visual grounding methods, which in can approximately recover our desired heatmap $\mathbf{H}$ by assembling unlimited crowdsourced proposals. We will leave the studies for this gap in the future. After training, 
% where $\mathcal{D}_o$ and $\mathcal{D}_e$ are auto- and human-annotated datasets in IVM-Mix-1M that contain image-instruction-mask pairs $(\mathbf{x}_{\rm img}, \mathbf{x}_{\rm txt}, \mathbf{y})$, respectively. The discriminator 

\textbf{Discriminator-weighted IVM Training}. Then, we apply the trained discriminator as a weighting function for the IVM training objective:
\begin{align}
    \min_\theta&\mathbb{E}_{(\mathbf{x}_{\rm img}, \mathbf{x}_{\rm txt}, \mathbf{H})\sim \mathcal{D}_o\cup\mathcal{D}_e}\left[f\left(d(\mathbf{x}_{\rm img}, \mathbf{x}_{\rm txt}, \mathbf{H})\right)\mathcal{L}^{\rm IVM}_{\theta}(\mathbf{x}_{\rm img}, \mathbf{x}_{\rm txt}, \mathbf{H})\right],\label{equ:ivm_train}\\
    &\mathcal{L}^{\rm IVM}_{\theta}(\mathbf{x}_{\rm img}, \mathbf{x}_{\rm txt}, \mathbf{H}) = \lambda_{\rm bce} \mathbf{BCE}(\hat{\mathbf{H}}_{\theta}, \mathbf{H})+\lambda_{\rm dice} \mathbf{DICE}(\hat{\mathbf{H}}_{\theta}, \mathbf{H}),
\end{align}
where $\lambda_{\rm bce}$ and $\lambda_{\rm dice}$ are set to 1.0 and 1.0 to balance the binary cross-entropy loss ($\mathbf{BCE}$) and the DICE loss for segmentation ($\mathbf{DICE}$)~\cite{jadon2020survey}, respectively. $f(x)\ge 0$ can be any non-negative, non-decreasing function. For simplicity, we set $f(x):=\min(\max(0.1, x), 1)$. This allows the weighting function $f(d(\cdot))$ in Eq.~(\ref{equ:ivm_train}) to dynamically prioritize training with high-quality data determined by the discriminator $d$. This approach maximizes the usage of reliable annotations in $\mathcal{D}_o$ to compensate for the small $\mathcal{D}_e$, while minimizing the impact of low-quality data in $\mathcal{D}_o$, thus optimizing performance.
% \zh{add brief explanation of BCE and DICE}

\subsection{Model Architecture}
The overall model framework is illustrated in Figure~\ref{fig: architecture}. Due to its complexity, IVM requires both reasoning and precise localization of the target object, closely paralleling reasoning segmentation~\cite{lai2024lisa}. Consequently, for the heatmap generator, we adopt a model design similar to that of LISA~\cite{lai2024lisa}. Specifically, we first extract dense image features using an isolated vision backbone and multimodal representation from an LMM, which processes image-instruction pairs. These two types of features are then fed into a lightweight generator that integrates them to produce a dense prediction.

% \begin{wrapfigure}{r}{6cm}
% \centering
% \includegraphics[width=0.4\textwidth]{Styles/Assets/architecture.png}
% \caption{\small architecture\ljx{to be updated}\zh{may be put a larger figure?}}
% \label{fig: architecture}
% \end{wrapfigure}
% the fig to be revised

For the discriminator, we deploy a lightweight discriminator that encodes the segmentation map using a two-layer convolution network. This discriminator interacts with the outputs of the LMMs through multiple cross-attention operators and finally outputs a quality score for each sample.

\textbf{Trainable Parameters}.
To enhance training efficiency, we freeze the pre-trained large foundation models and perform LoRA finetuning~\cite{hu2022LoRA}. The vision backbone, inherited from \textit{Segment Anything Model}~\cite{kirillov2023segment}, is completely frozen, while the lightweight generator and discriminator are fully finetuned. Notably, we utilize a shared LMM for both the generator and discriminator branches but employing separate LoRA parameters to avoid interference between the two tasks.
% \subsection{Analysis}
% \vspace{-5pt}

\section{Experiments}
\begin{figure}[h]
    \centering
    \includegraphics[width=0.99\linewidth]{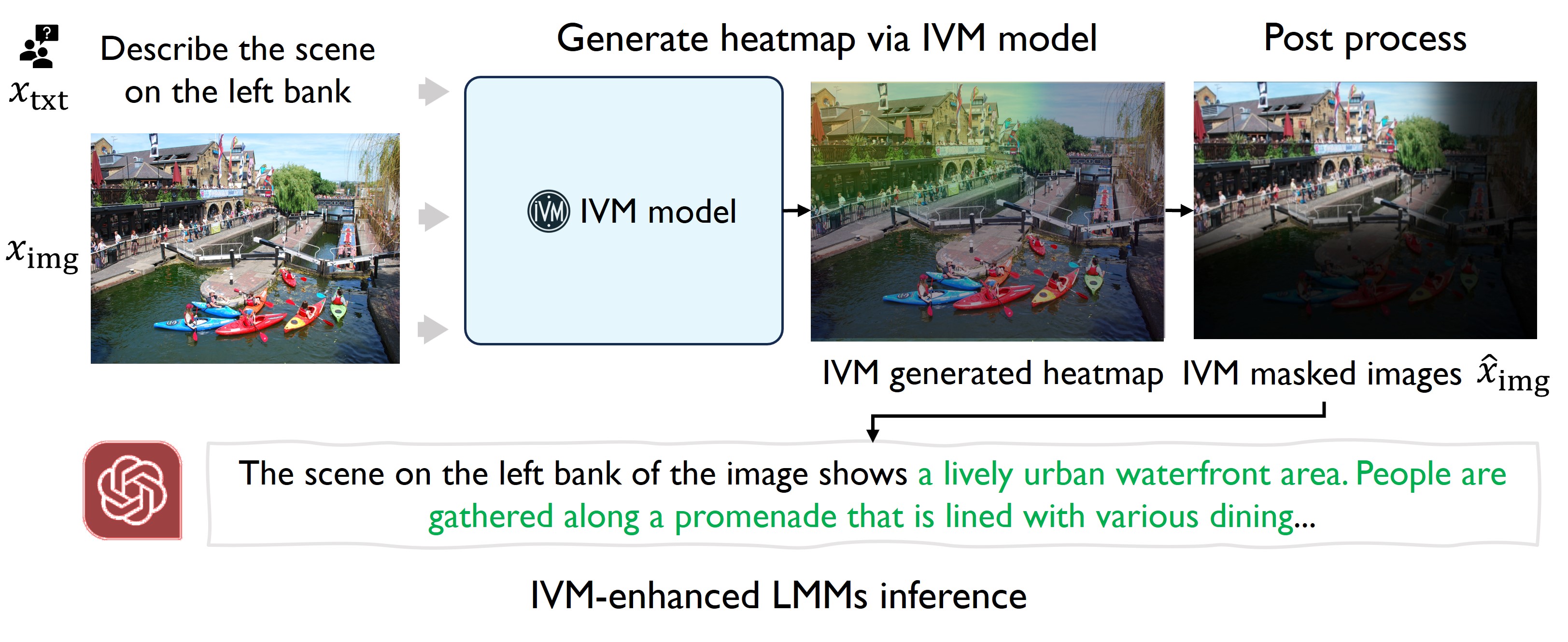}
    % \vspace{-5pt}
    \caption{\textbf{IVM inference pipeline.} IVM generates heatmap given a pair of image and instruction. Then, instruction-irrelevant visual areas are masked out via post process methods. LMMs can correctly follow the instruction given the masked images.}
    \label{fig:inference_pipeline}
\end{figure}

% \vspace{-5pt}
In this work, we employ \textit{LLaVA-7B}~\cite{liu2024visual} as the LMM and \textit{SAM-H}~\cite{kirillov2023segment} as the vision backbone for our IVM model (Figure~\ref{fig: architecture}), which is trained on the IVM-Mix-1M dataset using the proposed DWSL algorithm. 
More details on the architecture and training can be found in {Appendix~\ref{sec:train_eval_appenx}}.
We conduct extensive experiments to assess the effectiveness of the IVM model. 
Specifically, we utilize the heatmap generated by the IVM for image post-processing. These processed images can then be seamlessly fed into downstream multimodal models for diverse tasks, as shown in Figure~\ref{fig:inference_pipeline}. Unless otherwise specified, we use the image post-processing method of overlaying and cropping to discard instruction-irrelevant image content. A detailed discussion on post-processing methods is presented in Section~\ref{post-process}. We also provide more evaluation results and analysis in  Appendix~\ref{sec:more_results}. 
% \jj{Edited till here} 

\begin{figure*}[h]
% \vspace{-5pt}
    \begin{minipage}[h]{.52\textwidth}
    \scriptsize
    \centering
    % \caption{\small }
    \setlength{\tabcolsep}{4pt}
    \captionof{table}{\small V* bench results. 
    % \jj{Shrink font to the same as Table 2}\ljx{OK!}
    }
     % \jj{Where does the figure on the right come from? It's confusing. The caption also doesn't say what the figure is for. What does it mean 'keeps changed images'?}
    % \resizebox{0.7\linewidth}{!}{
    % \footnotesize
    \begin{tabular}{lccc}
        \toprule 

        \textbf{LMMs} & Attribute(\%) &Spatial(\%) & Overall(\%) \\

        \midrule
        
        \multicolumn{4}{c}{\textit{Open-Sourced LMMs}} \vspace{+4pt}\\
        BLIP2~\cite{li2023blip} & 27.0 & 53.9 & 37.7 \\
        MiniGPT-4~\cite{zhu2023minigpt} & 30.4& 50.0& 38.2 \\
        InstructBLIP~\cite{Dai2023InstructBLIP} & 25.2& 47.4& 34.0\\
        Otter~\cite{li2023otter} & 27.0& 56.6& 38.7\\
        LLaVA-1.5~\cite{llava_1_5} & 43.5 &56.6 &48.7\\

         \midrule
        \multicolumn{4}{c}{\textit{Commercial Chatbots}} \vspace{+4pt}\\
        Bard~\cite{manyika2023overview}& 31.3 & 46.1 & 37.2 \\
        Gemini-Pro~\cite{team2023gemini} & 40.9 & 59.2 & 48.2\\
        GPT4-V~\cite{gpt4} &   51.3 & 60.5 & 55.0 \\
        \midrule

        \multicolumn{4}{c}{\textit{Specific Visual Search Models}} \vspace{+4pt}\\
        SEAL~\cite{wu2023vstar} & 74.8~(+23.5) & 76.3~(+15.8) & 75.4~(+20.4) \\
        \midrule
        IVM-Enhanced GPT4-V & \textbf{87.0~(+35.7)} &
        \textbf{72.4~(+11.9)} & \textbf{81.2~(+26.2)}\\

        \bottomrule
    \end{tabular} 
    \label{tab:sota}
    \end{minipage}
    \hfill
    \begin{minipage}[h]{.45\textwidth}
        \vspace{-2pt}
        \centering
        \includegraphics[width=0.9\textwidth]{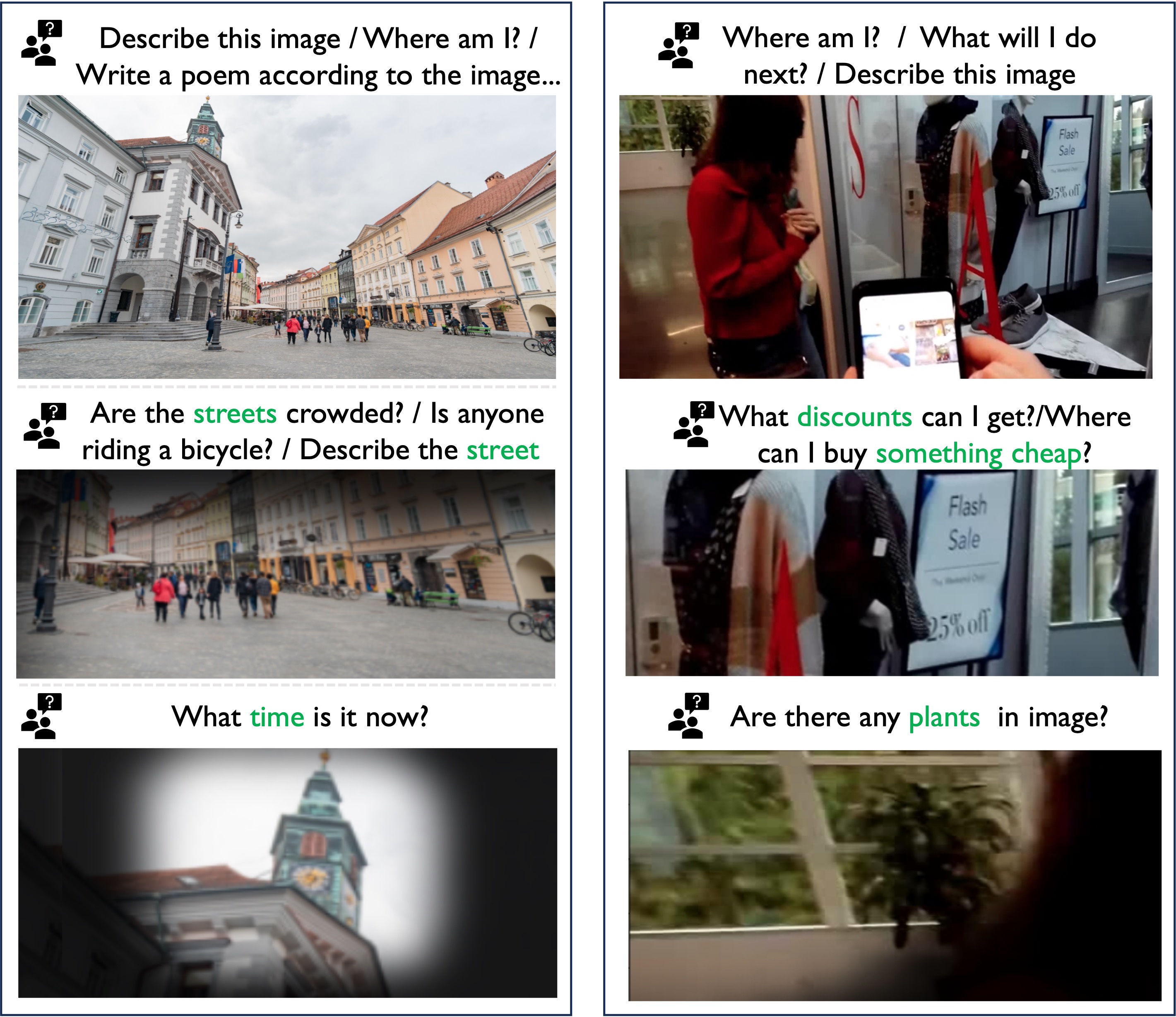}
        \captionof{figure}{\small
        % Visualizations of IVM-masked Images. 
        IVM can handle various instructions, ranging from retaining entire images for captioning (row 1) to localizing unique objects (row 2 and 3).
        % \jj{What's the difference between 1st image and the other two?}\ljx{the images below the instructions are IVM-masked instruction-related image regions} \jj{So the 1st image means no masking? Why include it?}\ljx{because these instructions are caption-style, and should use all image areas to answer the question}\jj{Perhaps should emphasize this in the caption, that the model knows when not to mask any region}\ljx{OK!}
        }
        \label{fig:enter-label}
    \end{minipage}
    \vspace{-0.5em}
\end{figure*}
% \vspace{-10pt}

\begin{table*}[t]
    \centering
    \vspace{-4pt}
    \caption{\small Results on other multimodal benchmarks.  MME* denotes the aggregate of scores from -p and -c.} 
    % \footnotesize
    \vspace{-5pt}
    \resizebox{1.0\linewidth}{!}{
    \begin{tabular}{lccccccc}
        \toprule 
        \textbf{LMMs} & \#Param & EgoThink &  POPE & MME* & GQA & SQA & VQAv2 \\
        \midrule
        InstructBLIP~\cite{Dai2023InstructBLIP} & 13B & - &78.9 & 1212.8  & 49.5 & 60.5 & -   \\
        Qwen-7B~\cite{bai2023qwen} & 7B & - & - & - & 58.3 & 67.1 & 78.8 \\
        SEAL-7B~\cite{wu2023vstar} & 7B &  - & 82.4 & 1129 & - & - & - \\
        LLaVA-7B~\cite{llava_1_5} &  7B &  51.1 & 85.9 & 1748 & 62.0 & 70.2 &78.5 \\
        LLaVA-13B~\cite{llava_1_5} & 13B & 55.2 &85.9& 1834 &67.1 & 71.6 & 80.0 \\
    
        \midrule

        LISA~\cite{lai2024lisa}-Enhanced LLaVA-7B & 20B & 47.9 (-3.2) & 80.0 (-5.9) & 1560 (-188) & 56.6 (-5.4) & 69.3 (-0.9) & 78.2 (-0.3)   \\
        \rowcolor[gray]{.9} 
        IVM-Enhanced LLaVA-7B & 14B & \textbf{54.5 (+3.4)} & \textbf{87.2 (+1.3)} & \textbf{1806 (+58)}   &  62.2 (+0.2) & 70.2 (-) & 79.0 (+0.5) \\ 
        
        \bottomrule
    \end{tabular} 
    }
    \label{tab:opensource}
    \vspace{-0.5em}
\end{table*}
\begin{figure}[t]
    \centering
    \hspace{-10pt}\includegraphics[width=1.0\textwidth]{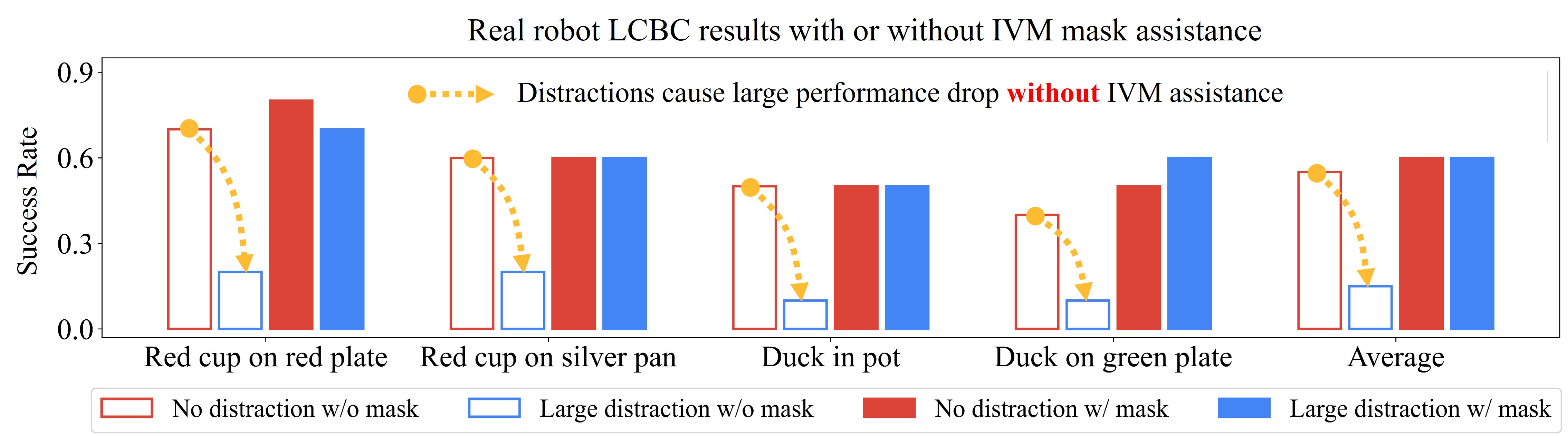}
    % \caption{\small}
    % \vspace{+5pt}
    \hspace{+5pt}
    \includegraphics[width=1.0\textwidth]{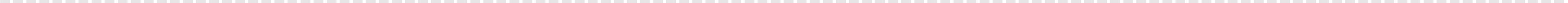}
    % \vspace{-5pt}
    \includegraphics[width=1.0\textwidth]{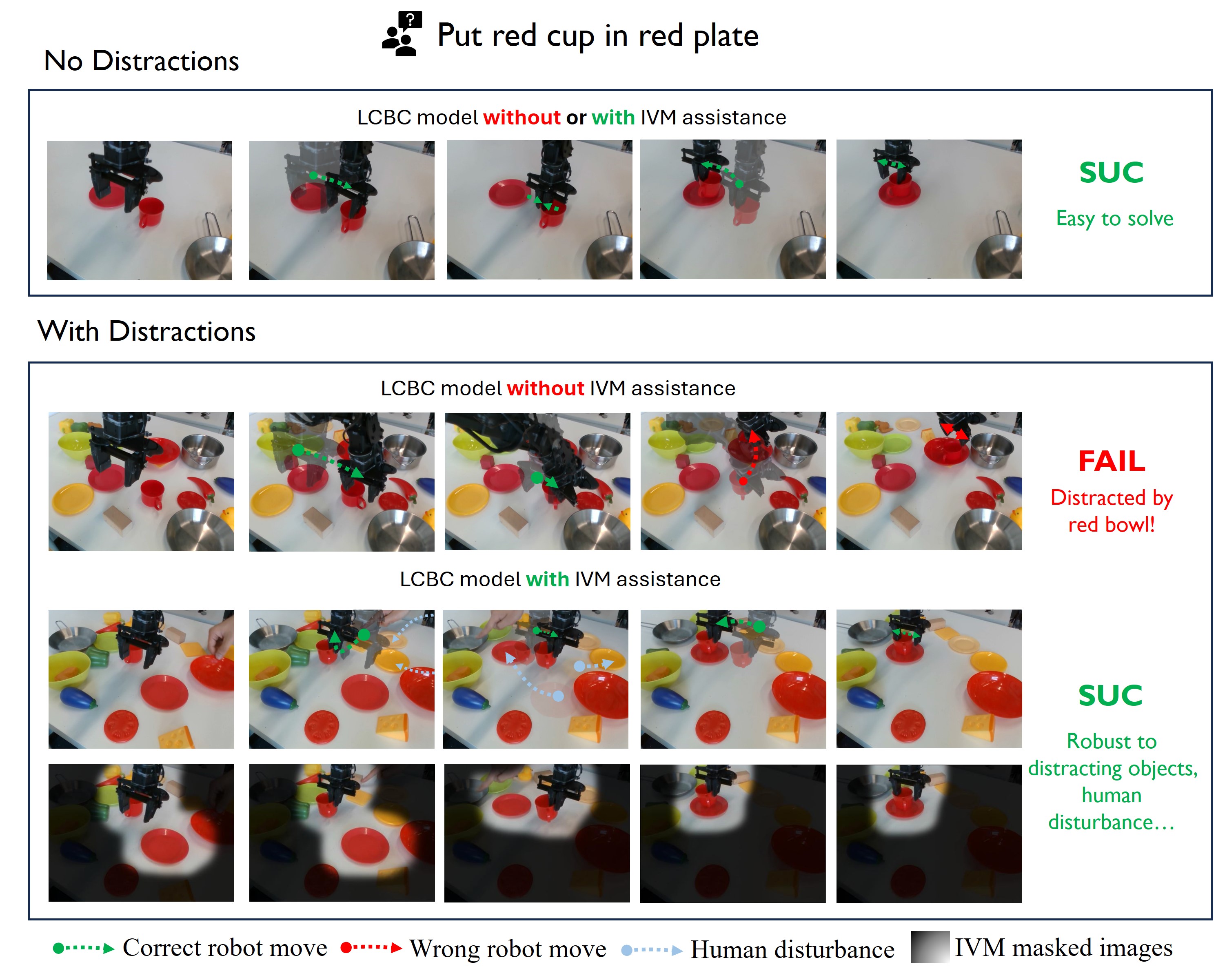}
    % \vspace{-15pt}
    \caption{\small Real robot results with or without IVM assistance. IVM greatly helps LCBC agent to overcome major
    % \jj{'Large distraction' is strange. Change 'large' to 'major' or 'big' in the figure}
    distractions, enjoying better robustness and generalization.
    % \jj{Typo in the figure: 'distract objects' to 'distracting objects'}
    See Appendix~\ref{subsec:real_robot_appenx} for experiment setups. 
    % \jj{'Big distraction' is still weird. How about just 'With distractions'?}\ljx{OK!} 
    % \jj{The green/red/blue arrows in the images are very hard to see. Perhaps enlarge them}
    }
    \label{fig:robot_main}
\end{figure}

\subsection{Main Results}

\textbf{Integration with Commercial Chatbot}. We use GPT4-V~\cite{gpt4} as the base model. Considering the superior perception and reasoning capability of  GPT4-V, we evaluate IVM-enhanced GPT4-V on V*bench~\cite{wu2023vstar}, a recently proposed challenging VQA-type benchmark  characterized by images with abundant redundancies. Results are presented in Table~\ref{tab:sota}. The accuracy of the vanilla GPT4-V is mediocre (55.0\%). Our IVM model, however, can significantly improve the performance (+26.2\%) and establish a new state of the art on this benchmark, even surpassing the task-specialized SEAL~\cite{wu2023vstar} that requires a complex heuristic visual search pipeline.

\textbf{Integration with Open-sourced LMMs}. To demonstrate the  versatility of our IVM model, we further integrate it into an open-sourced LMM, LLaVA-7B~\cite{llava_1_5}. We conduct extensive experiments across various benchmarks, including EgoThink~\cite{cheng2024egothink}, POPE~\cite{li2023pope}, MME~\cite{fu2023mme}, GQA~\cite{hudson2019gqa}, SQA~\cite{lu2022SQA}, and VQAv2~\cite{goyal2017vqav2}. As shown in Table~\ref{tab:opensource}, our IVM-enhanced LLava-7B gains consistent performance improvements, achieving comparable performance to (even surpassing) LLaVA-13B on EgoThink, POPE and MME. Although IVM-enhanced LLaVA-7B  and LLaVA-13B~\cite{llava_1_5} have roughly the same number of parameters, the latter integrates more powerful pretrained foundation models. In contrast, our IVM model allows the 7B model to outperform the 13B model by merely simplifying visual input, further validating the power of visual masking.

Meanwhile, IVM-enhanced LLaVA-7B does not show significant gains on GQA, SQA and VQAv2, which is expected, as these benchmarks do not heavily rely on grounding capabilities: VQAv2 and GQA contain relatively simple visual input where most regions of the images are instruction-relevant, while SQA primarily focuses on assessing model reasoning capability.

\textbf{Comparison with Reasoning Segmentation Model}. We also compare against LISA~\cite{lai2024lisa}, which is most analogous to IVM. We provide carefully tailored prompts like \textit{"what should we focus on the image to follow the given instruction? Give me the seg"} to extend LISA into visual masking task. However, even with larger 13B model and extensive tuning of input prompt, masks generated by LISA consistently result in severe performance degradation on all tasks.
%Except for the instruction, we design extra prompts for LISA to adapt it to our task: \textit{"what should we focus on the image to follow the given instruction. Give me the seg."} 

\textbf{Evaluation on Real Robotic Control}. We also plug the IVM model into robot control tasks to help robot model improve generalization. Specifically, we evaluate a language-conditioned behavior cloning (LCBC) robot agent trained with or without IVM masked images. Figure~\ref{fig:robot_main} clearly demonstrates that without IVM assistance, the LCBC robot agent suffers from severe performance drop when noticeable distractions are applied. With IVM assistance, however, the agent consistently pays close attention to correct instruction-related image regions, performing robustly against diverse distractions such as  human disturbances and numerous task-irrelevant objects of various colors and shapes. This demonstrates promising potentials of using IVM to enhance embodied agents to follow complex instructions in unseen scenes with plenty distractions.

% \vspace{-2pt}
\subsection{Ablation}
% \vspace{-2pt}
We ablate the key components of IVM and report overall accuracy improvement(\%) of IVM-Enhanced GPT4-V, evaluated on V* bench~\cite{wu2023vstar} due to its high demand on precise visual grounding abilities.

\begin{figure}[t]
\centering
\includegraphics[width=0.99\textwidth]{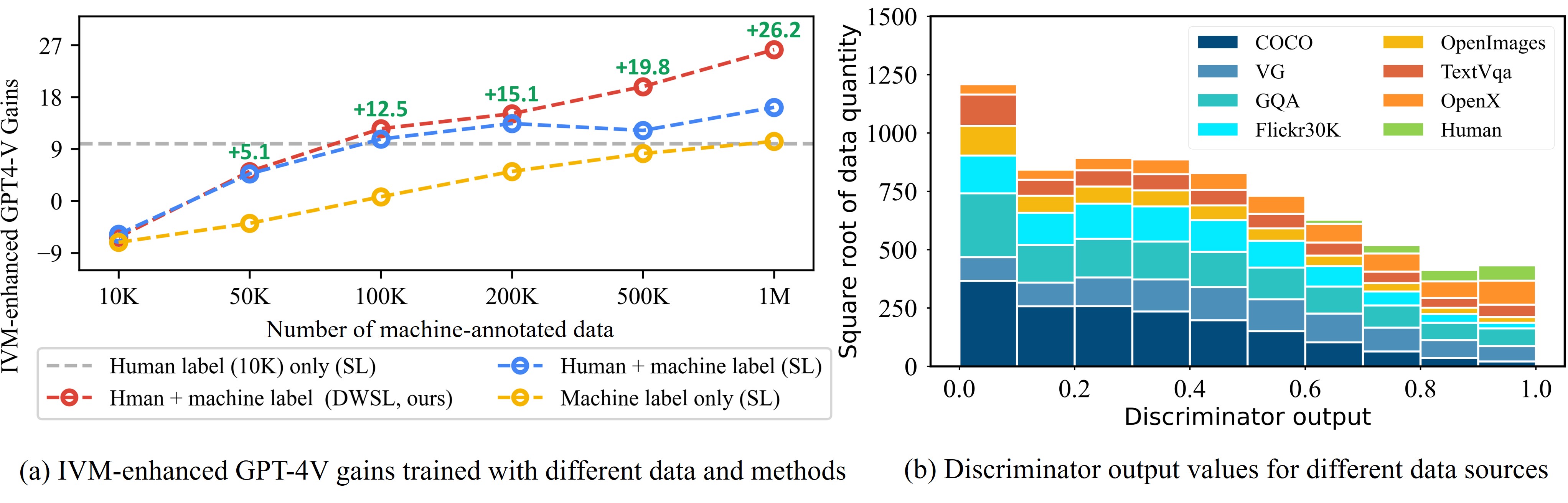}
\vspace{-5pt}
\caption{\small Ablations on training data and the proposed DWSL framework.
% \jj{In (a), it seems it's only about trained with different data, not different methods?}
}
\label{fig:dwsl_ablation}
% \vspace{-5pt}
\end{figure}

% \begin{table}[h]
%     \centering
%     \caption{Mask deployment strategy}
%     \vspace{+5pt}
%     \begin{tabular}{cccc}
%         \toprule
%         & overlay & blur & gray-scale  \\
%          \midrule
%         w/ crop & +26.2 & +24.4 & +22.1 \\
%         % \midrule
%         w/o & +19.1 & +17.2 & +10.2 \\
%         \bottomrule
%     \end{tabular}
%     \label{tab:mask}
% \end{table}

\begin{figure}[t]
\vspace{-5pt}
\hfill
\begin{minipage}[h]{.44\textwidth}
    \centering
    \includegraphics[width=0.99\textwidth]{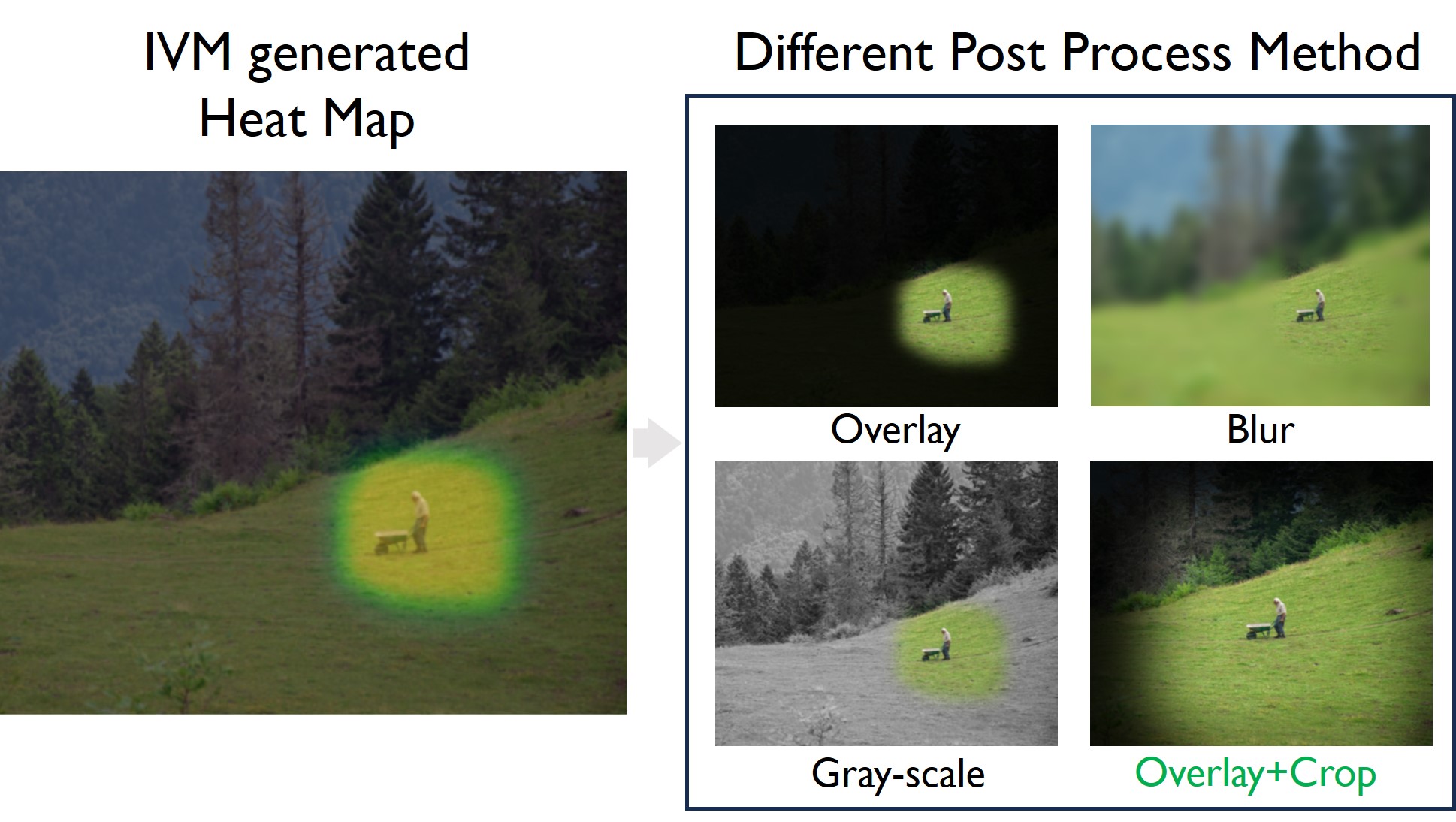}
    \caption{\small Different mask deployment methods.}
    % \vspace{-5pt}
    \label{fig:enter-label}    
\end{minipage}
\hspace{+2pt}
\hfill
\begin{minipage}[h]{.45\textwidth}
    \centering
    \small
    \captionof{table}{\small Ablations on different mask deployment methods on the V*bench.}
    \begin{tabular}{cccc}
        \toprule
        & Overlay & Blur & Gray-scale  \\
         \midrule
        w/ crop & \textbf{+26.2} & +24.4 & +22.1 \\
        % \midrule
        w/o & +19.1 & +17.2 & +10.2 \\
        \bottomrule
    \end{tabular}
    \label{tab:mask}
\end{minipage}
\hfill
\vspace{-5pt}
\end{figure}

\textbf{Training Data}. We investigate the impact of IVM-Mix data characteristics on IVM performance from two key perspectives: $1)$ Large machine-annotated data volume clearly enhances IVM model performance, as illustrated by the progressive improvement in Figure~\ref{fig:dwsl_ablation} (a) with increased machine-annotated data volume (red, blue and yellow line). This demonstrates the effectiveness of our proposed LLM-empowered Mixture-of-Expert pipeline in generating reliable data for IVM training. $2)$ Figure~\ref{fig:dwsl_ablation} (a) also reveals that incorporating human annotations significantly boosts training efficiency (red and blue \textit{v.s} yellow line), highlighting the critical role of introducing human preferences in IVM-Mix-1M dataset, despite its relatively small volume compared to machine-annotated data (only 1:100).
% $1)$ We randomly selected subsets of varying sizes from the IVM-Mix dataset and trained models on these subsets under the same settings to assess the impact of data volume on model performance. The results, depicted by the red lines in Figure~\ref{fig:dwsl_ablation} (a), indicate that the performance of our IVM model progressively improves with increased training data. $2)$ We separate the manually labeled data within the IVM-Mix dataset to examine the effect of incorporating human preferences into the training process.  The performances of the IVM model trained exclusively with machine-generated data and human data (10K) are illustrated by the yellow lines and gray lines separately, in Figure~\ref{fig:dwsl_ablation} (a). We observe that the model performance consistently improves with increased data even without human labels, but the training efficiency is quite limited. Eg. A model trained with 1M-machine-generated data can only achieve performance comparable to that obtained from 10K human data (100:1). 

\textbf{DWSL Framework}. We also explore the efficacy of the DWSL framework in Figure~\ref{fig:dwsl_ablation} (a) by comparing IVM training using: $1)$ DWSL (red line), $2)$ traditional Supervised Learning (SL) without DWSL (blue line), and $3)$ SL on limited human data (gray line). The results demonstrate that DWSL effectively leverages both human- and auto-annotated data, particularly as the volume of machine-annotated data increases, enjoying higher asymptotic performances. This is expected as machine-annotated data often contain inaccuracies and training naively using all these data can lead to suboptimal results. Meanwhile, the limited human data alone cannot provide satisfactory outcomes. DWSL, however, addresses these challenges by dynamically prioritizing good samples and discarding misleading ones, resulting in stable and improved results. This is further illustrated in Figure~\ref{fig:dwsl_ablation} (b) which visualizes the outputs of the discriminator for each sample, where the discriminator can correctly retain good samples (e.g. Human) and filter out low-quality data with lower weights. 

% drop the discriminator in IVM and train it with vanilla supervised learning (blue line in Figure~\ref{fig:dwsl_ablation} (a)).
% We report the result with the blue line in Figure~\ref{fig:dwsl_ablation} (a)
% By comparing the result to our DWSL-trained model on the red line, we can get the following conclusion: The DWSL algorithm does not exhibit significant advantages with smaller datasets(Eg. <100K). However, as the data volume increases, so does the proportion of noise from machine-generated labels, which can diminish learning efficiency and even lead to a decline in performance. In such scenarios, the DWSL algorithm becomes particularly valuable. It effectively guides the model to focus on learning from high-quality data, achieving more stable and improved performance outcomes.

% To further illustrate this,
% Furthermore, to deepen the understanding of our DWSL algorithm, 
% we visualize the output weights of the discriminator for each sample in the IVM-Mix dataset in Figure~\ref{fig:dwsl_ablation} (b). This visualization reveals that a substantial portion of low-quality data (with weights < 0.1) was effectively filtered out while partial high-quality machine-generated data was accurately retained by the discriminator.

\textbf{Mask Deployment Strategy}. We investigate the impact of mask deployment strategy on downstream applications. While more complex solutions such as visual search algorithms~\cite{wu2023vstar} can be employed, our investigation focuses solely on simpler approaches to understand the intrinsic capabilities of IVM model. Specifically, we examine four basic masking methods: overlay, blur, grayscale, and cropping, as illustrated in Figure~\ref{fig:enter-label}. In particular, for the crop method, we find the smallest area that retains all the activated (>0) values in the heatmap and crop it. Table~\ref{tab:mask} demonstrates that IVM maintains robustness across all simple post-processing methods, where overlay+crop enjoys the most performance enhancement and thus is used as our default mask deployment method.
\label{post-process}

% \vspace{-3pt}
\section{Conclusion}
% \vspace{-3pt}
We introduce Instruction-guided Visual Masking (IVM), a generic and powerful visual grounding method that enhances broad multimodal instruction following tasks in a plug-and-play way.
By masking out all instruction-irrelevant image regions, IVM effectively injects superior visual grounding ability to downstream LMMs non-intrusively, significantly boosting both commercial and open-sourced LMMs and achieving state-of-the-art results across numerous challenging multimodal benchmarks. Real robot experiments further demonstrate the versatility of IVM, showcasing the potential to deploy IVM to embodied robotic tasks where failures caused by distractions are long-standing challenges. For further improvement, one promising direction is to finetune LMMs using IVM-generated heatmap as an additional input channel to reduce suboptimal heuristics caused by mask deployment methods. Due to resource limitation, we leave this for future work. We open source the IVM checkpoint and the IVM-Mix-1M dataset to help the community further explore relevant directions\footnote{\href{https://github.com/2toinf/IVM}{\texttt{https://github.com/2toinf/IVM}}}. More discussion on limitations and future directions can be found in Appendix~\ref{sec:limitation}.

% In this paper, we introduce Instruction-guided Visual Masking (IVM), a generic and powerful visual grounding method, that is versatile to enhance diverse multimodal instruction following tasks such as VQA, visual captioning and robotics control. 
% % a generic and powerful visual grounding method, that can effectively enhance diverse multimodal instruction following tasks by directly injecting sup
% Serving as a plug-and-play tool, IVM effectively injects superior visual grounding ability to downstream multimodal models. By simply masking out all instruction-irrelevant image regions,
% % to zoom in instruction-sensitive areas, 
% IVM can help both commercial and open-sourced LMMs reach new state-of-the-art results across numerous challenging multimodal benchmarks 
% % such as V*bench, EgoThink, POPE, MME 
% and greatly enhances generalizations for embodied robot agent. This demonstrates promising potential to extend IVM on more challenging multimodal instruction following tasks, \textit{e.g.}, . To help the community further explore this, we will release our IVM model as well as all the IVM-Mix-1M data upon acceptance. 
% % Although the plug-and-play paradigm may require some heuristic post-processing designs, we systematically
% Discussion on limitations and future directions can be found in Appendix ~\ref{sec:limitation}.

% We also introduce the IVM-Mix-1M, a high-quality visual grounding dataset, that contains one million data samples attached with complex instructions and annotations. 

% significantly enhancing the performances of existing LMMs and embodied robot model 

\section{Acknowledgements}
The paper is supported by funding from Wuxi Research Institute of Applied Technologies, Tsinghua University under Grant 20242001120. The authors would like to thank the anonymous reviewers for their feedback on the
manuscripts.

\bibliographystyle{named}
\bibliography{neurips_2024}

\newpage
\appendix
% \section{Appendix / supplemental material}
% Optionally include supplemental material (complete proofs, additional experiments and plots) in appendix.
% All such materials \textbf{SHOULD be included in the main submission.}

\section{Limitation and Future Work}
\label{sec:limitation}

Here, we discuss our limitations, potential solutions and interesting future works.
\begin{enumerate}
    \item \textbf{Computational Overhead}. Note that IVM introduces additional parameters and computational overhead to directly enhance visual grounding ability of LMMs, which in turn indirectly improve the VQA performances. However, more VQA performance gains can be obtained if the same amount of additional parameters are end-to-end trained directly on VQA data (LLaVA-13B v.s IVM-Enhanced LLaVa-7B in Table~\ref{tab:opensource}). 

    \textit{Solution and future work:} Nevertheless, this is quite reasonable because IVM primarily focuses on improving the visual grounding ability, but accurate VQA also requires other abilities which can be learned through end-to-end training. End-to-end training, however, requires tremendous VQA data to implicitly and slowly improve the visual grounding ability, which is quite data-intensive. Both Table~\ref{tab:sota} and Figure~\ref{fig:lmm_fail} can show that even trained on billions of data, GPT4-V still performs subpar on tasks that require strong visual grounding ability. IVM, instead, can significantly boost the visual grounding ability of GPT4-V using just 7B parameters and less computations. One promising and interesting future direction is to include some auxiliary tasks to directly absorb the strong visual grounding ability in the IVM-Mix-1M dataset through end-to-end training like~\cite{wu2023see}.

    \item \textbf{Data Quality}. Due to task complexity, the machine-annotated data in IVM-Mix-1M inevitably includes wrong labels that mistakenly exclude instruction-sensitive image regions or suboptimal labels that not fully mask out all instruction-irrelevant areas. These inaccuracies may lead to suboptimal IVM model. We propose a DWSL framework to tackle this. However, the DWSL framework relies on a learned discriminator and a human-designed $f(x)$ function, which may not exclude all inaccuracies.

    \textit{Solution and future work:} We have clearly demonstrated in Figure~\ref{fig:dwsl_ablation} that with a simple $f(x)$ and a lightweight discriminator, DWSL consistently outperforms the naive Supervised Learning (SL), doing pretty well on prioritizing good samples and meanwhile identifying inaccurate labels. To further enhance this, one can use other advanced techniques such as Reinforcement Learning from Human Feedback (RLHF)~\cite{ouyang2022training, bai2022training, hu2024querypolicy} to provide more fine-grained judgement on annotation qualities, or resort a theoretical-soundness $f(x)$~\cite{xu2022discriminator} to achieve better results. In addition, one can also use our pretrained IVM model to directly generate high-quality heatmaps to enhance the machine annotations.
    
    \item \textbf{Mask Deployment Methods}. In this paper, we directly use the simple post-processing method to apply the IVM generated heatmaps on images, which then are fed into LMMs to perform downstream tasks. However, these post-processing methods introduce some heuristics, which may be suboptimal for downstream LMMs. In addition, LMMs may not see many masked images during pretraining, thus some distributional shift may occur.

    \textit{Solution and future work:} Although these limitations exist, IVM still obtain consistent improvements using diverse mask deployment methods, as shown in Table~\ref{tab:mask}, which showcases the great versatility of IVM to inject visual grounding abilities. To further improve this, one strategy is employ some task-specialized visual search method~\cite{wu2023vstar}, but will bring many computational load during inference and limit the versatility on embodied agents. Another promising direction is directly using the IVM generated heatmaps as an additional input channel to finetune the LMMs like~\cite{sun2023alphaclip}, which can fully eliminate the heuristics of post-process methods, may bring larger performance gains. Due to resources limits, we leave this for a future work.

    \item \textbf{Fine-grained Heatmaps}. Note that the IVM generated heatmaps cannot provide exact semantic object segmentation with clear contours like reasoning segmentation~\cite{lai2024lisa} offers. 

    \textit{Discussions:} We want to clarify that this is an advantage of the IVM model rather than a limitation. This is because of the ambiguous nature of the visual masking task. For this task, the ground truth heatmaps are mostly less semantic-meaningful for annotations as discussed in~\ref{subsec:problem_def}. So, we ensemble the annotation proposals from different visual grounding methods for data annotation, which will make the trained IVM model robust to include instruction-relevant image areas, rather than being aggressive to exclude some instruction-sensitive pixels like reasoning segmentation~\cite{lai2024lisa} does illustrated in Figure~\ref{fig:intro}.
    
\end{enumerate}

Overall, although some limitation exist, we have thoroughly discussed potential solutions to these limitations. Moreover, in this paper, we have demonstrated the superior effectiveness and versatility of IVM to directly inject strong visual grounding ability to downstream LMMs or embodied agents, representing a pioneer effort to extend traditional visual grounding methods towards a more complex and generic setting that covers diverse multimodal instruction following tasks.

\section{Broader Impact}
\label{sec:broad_impact}
This paper aims to advance the field of artificial intelligence, where no significant negative social impact is observed in this paper. The IVM-Mix-1M may contain some potential privacy issues and biases. However, in this paper, nearly all data are collected from open-sourced data, which have been well peer-reviewed, thus resolved this ethical concern.

\section{Training and Evaluation Details}
\label{sec:train_eval_appenx}

\subsection{Architecture Details}
\label{subsec:arch_appenx}
\begin{figure}[h]
    \centering
    \includegraphics[width=0.8\textwidth]
    {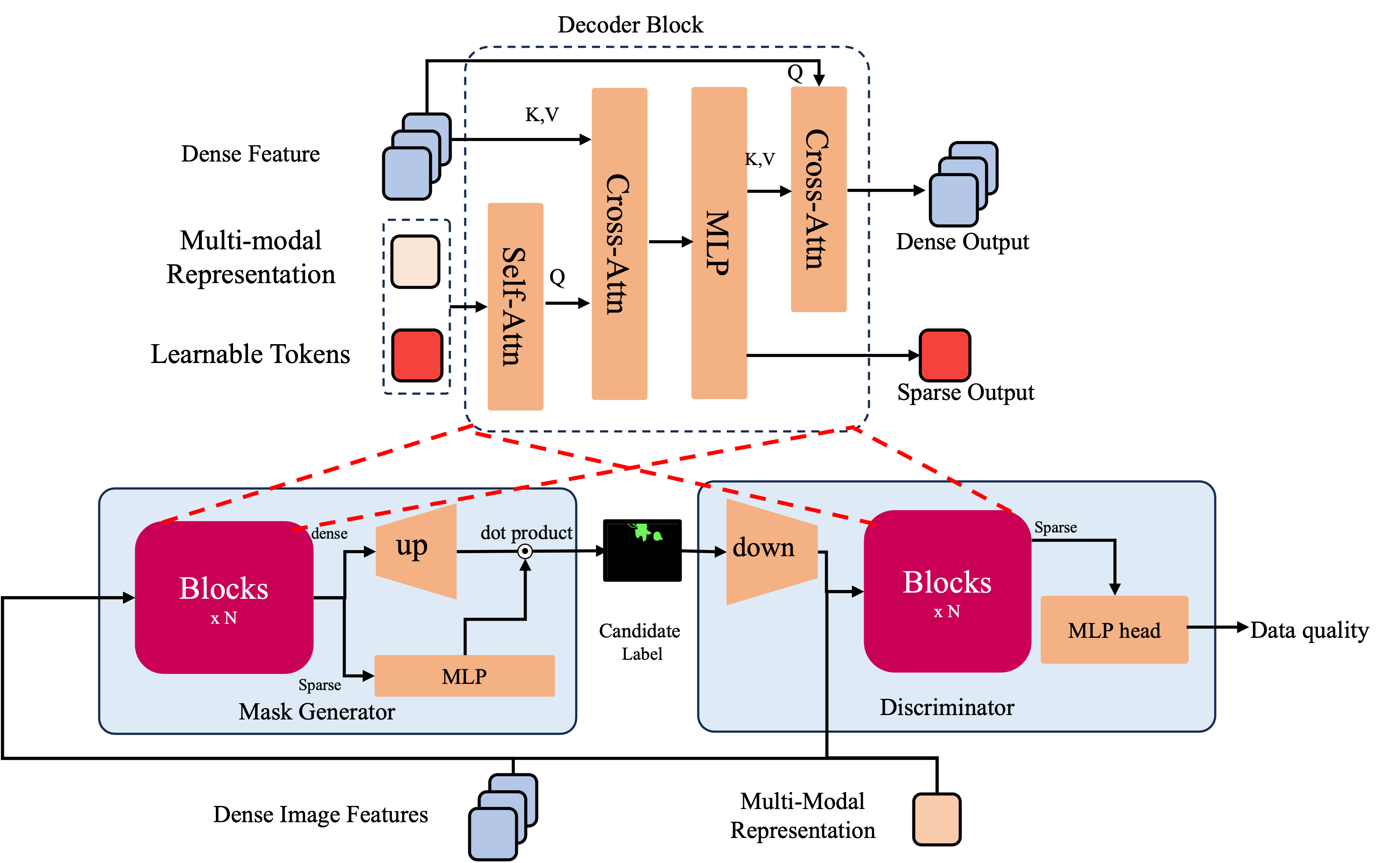}
    \caption{Generator/Discriminator Architecture Details}
    \label{fig:decoder}
\end{figure}

In this section, we primarily focus on the architectural design of the lightweight generator and discriminator, as both the Language Model Multitask (LMM) and the vision backbone are derived from the powerful foundation models (LLaVA \& SAM). Both the generator and discriminator utilize the same transformer-based decoder block, as depicted in Figure~\ref{fig:decoder}. We employ two such blocks for both the generator and discriminator. Specifically, the generator produces dense predictions by upscaling the output features of the decoder block through a straightforward upsampling operation. In contrast, the discriminator first employs a two-layer convolutional downsampling network to encode segmentation labels. This network, in conjunction with the decoder block and a simple MLP (multi-layer perceptron) head, outputs the weights.

\subsection{Training Details}
\label{subsec:train_appenx}

We adopt 8 NVIDIA 80G A100 GPUs and take 4 days to train our IVM model. The training scripts
are based on deepspeed~\cite{aminabadi2022deepspeed} engine and the training hyperparameters can be found in Table~\ref{tab:train-setting}.
% The weights of discriminator loss and the mask generator loss loss are set to 0.01 and 1.0, respectively,
% and those of the bce loss and the dice loss are set to 1.0 and 1.0, respectively.

\begin{table}[h]
\caption{Hyper-parameters for pretraining.}
\vspace{+5pt}
    \centering
    % \resizebox{0.4\textwidth}{!}{
    \begin{tabular}{lc}
        \toprule
        config & value \\
        \midrule
        training iteration & 200K  \\
        optimizer & AdamW~\cite{adamw} \\
        learning rate & $1\times10^{-5}$ \\
        batch size & 32 \\
        weight decay & 0 \\
        optimizer momentum & $\beta_1,\beta_2$=0.9, 0.95  \\
        data augmentation & \textit{RandomCropResize} \\
        \bottomrule
    \end{tabular}
    % }
    \label{tab:train-setting}
\end{table}

\subsection{Multimodal Benchmarks Evaluations}
\label{subsec:mmbench_appenx}
We evaluate our IVM on diverse multimodal benchmarks, including general VQA (VQAv2~\cite{goyal2017vqav2}, GQA~\cite{hudson2019gqa}, MME~\cite{fu2023mme}),  first-person perspective QA (EgoThink~\cite{cheng2024egothink}), scientific QA (SQA~\cite{lu2022SQA}),  hallucination adversarial QA (POPE~\cite{li2023pope}) and V*~\cite{wu2023vstar}, a recently proposed challenging benchmark with high-resolution and complex visual input.

Our evaluation employs a two-stage inference pipeline: the image is firstly simplified by IVM-generated heatmap and mask deployment methods; Subsequently, the simplified image is fed into downstream LMMs(GPT4-V~\cite{gpt4}, LLaVA~\cite{llava_1_5}) without finetuning. We adhere to the official procedures of each benchmark to evaluate the output of LMMs and report the results.

\subsection{Real Robot Evaluations}
\label{subsec:real_robot_appenx}

\begin{figure}[h]
    \centering
    \includegraphics[width=0.75\textwidth]{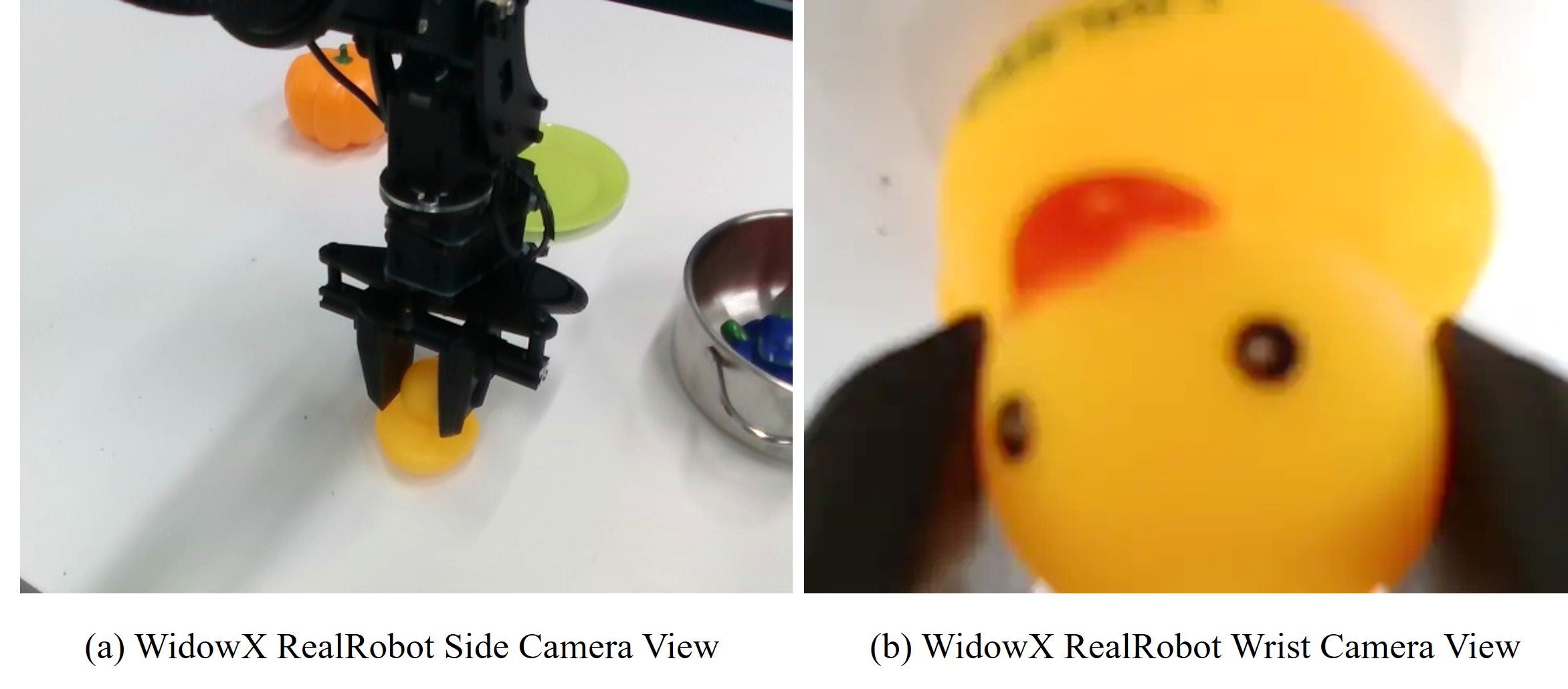}
    \caption{Visual input view for LCBC policy.}
    \label{fig:view}
\end{figure}

\textbf{Task descriptions}. The real robot experiments evaluate several \texttt{pick and place} manipulation tasks that require strong visual grounding abilities. Specifically, we evaluate on 4 tasks as shown in Table~\ref{tab:real_robot}, following the task definitions in DecisionNCE~\cite{li2024decisionnce}. For each task, we collect around 100 demonstrations using the
demonstration collection system in BridgedataV2~\cite{walke2023bridgedata}. We take both a side camera view and a wrist camera view as the vision inputs, as shown in Figure~\ref{fig:view}. For each demonstration, the environmental steps are
around 50 steps. During data collection, the object and robot locations are randomly initialized, and the scene also has lots of randomly located
distractors with varied shape and color. 

\begin{table}[h]
    \centering
    \caption{Real Robot Tasks}
    \begin{tabular}{ll}
    \toprule
     Environment ID  & Language Instruction \\
     \midrule
     \texttt{Red cup on silver pan} & \texttt{Pick up the red cup and place it on the silver pan}\\
     \texttt{Red cup on red plate} & \texttt{Pick up the red cup and place it on the red plate}\\
     \texttt{Duck on green plate} & \texttt{Pick up the duck and place it on the green plate}\\
     \texttt{Duck in pot} & \texttt{Pick up the duck and place it in the pot}\\
     \bottomrule
    \end{tabular}
    \label{tab:real_robot}
\end{table}

\textbf{Training details}. Here, we train Language-Conditioned Behavior Cloning (LCBC) policies using DDPM~\cite{ho2020denoising} loss since diffusion policies are good at fitting complex data distributions~\cite{zheng2024safe, walke2023bridgedata, ajay2022conditional}, especially human demonstrations~\cite{chi2023diffusion}. For model architecture, the side and wrist images are augmented and then passed through a shared ResNet50~\cite{he2016deep} image encoder and get an image embedding for each camera view, following~\cite{walke2023bridgedata}. As the downstream data is quite limited, we load the ImageNet~\cite{deng2009imagenet}-pretrained ResNet50 image encoder and further train it on the small robot data. Meanwhile, the language instruction is passed through a frozen T5 text-encoder~\cite{raffel2020exploring}, which is fused into the image encoder via Film conditioning layers~\cite{perez2018film}. Then, this language-conditioned image embedding is passed through a MLPs with residual connections similar to IDQL~\cite{hansen2023idql}, which then outputs the predicted noise in DDPM~\cite{ho2020denoising}. To obtain smoothed policy rollouts, we adopt Action Chunking and Temporal Ensemble from ACT~\cite{zhao2023learning} with a chunking size 4 rather than 100 in~\cite{zhao2023learning} because the episode horizons in this paper are only around 50. The LCBC policies are trained either on the original side camera view (without IVM assistance) or on the IVM-masked side camera view (with IVM assistance) for 200K steps with a batch size of 64. The training can be completed on 2 NVIDIA RTX4090 GPU in 17h. All hyperparameters are summarized in Table~\ref{tab:robot_exp}.

\begin{table}[h]
    \centering
    \caption{Real robot LCBC training details}
    \begin{tabular}{lc}
    \toprule
     \multicolumn{2}{c}{Backbones}\\
     \midrule
     Visual encoder & Resnet50~\cite{he2016deep} (ImageNet~\cite{deng2009imagenet} pretrained) \\
     text encoder & T5~\cite{raffel2020exploring} (frozen) \\
    % \bottomrule

     \toprule
     \multicolumn{2}{c}{DDPM hyperparameters}\vspace{+0pt}\\
     \midrule
     noise schedule & VP~\cite{song2021score} \\
     denoising time steps & 25 \\
     % \bottomrule

     \toprule
     \multicolumn{2}{c}{Other hyperparameters}\vspace{+0pt}\\
     \midrule
     Chunking size & 4 \\
     Optimizer  & AdamW~\cite{adamw} \\
     Learning rate & 1e-4 \\
     Lr schedule & cosine annealing \\
     Warm up steps & 2000 \\
     Batch size & 64 \\
     Gradient Steps & 200K \\
     Augmentation & Yes~\cite{walke2023bridgedata} \\

    %  \toprule
    %  \multicolumn{2}{c}{Finetuning hyperparameters}\vspace{+0pt}\\
    %  \midrule
    %  Learning rate & 1e-5 \\
    %  Batch size & 64 \\
    %  Gradient Steps & 100K \\
    %  Augmentation & Yes~\cite{walke2023bridgedata} \\
    \bottomrule
    \end{tabular}
    \label{tab:robot_exp}
\end{table}

\textbf{Evaluation details}. We first evaluate the trained LCBC policies without strong distractions, where no or only small distractors appear in the image. Then, we add lots of distracting objects with varied shapes and colors, and even introduce strong human disturbance to attack the LCBC policies. For each score reported in Figure~\ref{fig:robot_main}, we evaluate 10 episodes and report the success rates.

\section{Mixture of Expert Annotation Pipeline}

\subsection{Labeled Visual Grounding data}

For labeled visual grounding data, We provide the following prompt to drive GPT-4~\cite{gpt4} to generate more complex instructions based on given language annotations.

\begin{mdframed}[style=boxstyle]

[Image Description]
\%s

[System]
You are an AI visual assistant, and you are seeing a single image. What you see are part of the image and are provided with a simple phrase.  Please generate any instructions that can be executed based on the content of the picture described, including simple queries about the content of the picture, such as the object types, counting the objects, object actions, relative positions between objects, etc. Also consider more complex questions that require reasoning. For example, you can ask what time it is now for a clock and what can I use to clean the room for a broom. Ensure that the questions you ask can be clearly answered only based on what you see. Please generate as many five questions as possible and return them in a single line separated by ';' and avoid any other output.

\end{mdframed}

\subsection{Unlabeled Visual Instruction Following Data}

For unlabeled visual instruction following data, we first try to simplify complex instructions.  Specifically, we employ GPT-4 to infer the necessary object for executing the given instructions based on these instructions and a simple image caption. If the dataset lacks captions, they can be generated using an existing caption model like BLIP-Caption~\cite{li2023blip}. Below, we outline the prompts specifically designed for GPT-4.

\begin{mdframed}[style=boxstyle]
[Image Caption] \%s

[Instruction] \%s

[System] 
You are an helpful AI assistant. I need to reply to the previous instruction based on an image, and I have a simple caption for the image. Please note that there may be objects in the image that I did not detect. Since you cannot view the image, please list any potential objects that might influence my responses, separated by semicolons, in a single line without any additional output. If you believe that the number of objects could be too extensive and might hinder my judgment, print 'None'.

\end{mdframed}
With the simplified instruction, we can adopt existing visual grounding models to generate the candidate label. Specifically, we utilize four models: AlphaCLIP~\cite{sun2023alphaclip}, LISA~\cite{lai2024lisa}, OwVIT~\cite{owdetr} and Grounding-SAM~\cite{ren2024grounded} and the inference pipelines are provided in the official implementation of these models.

\section{More result}
\label{sec:more_results}

\subsection{Referring Expression Comprehension}

As IVM is an extension of traditional visual grounding task, we also evaluate our IVM on RefCoCo, RefCoCo+ and RefCoCog~\cite{yu2016rec}. We reported the accuracy (IOU-50\%) on the validation split in Table~\ref{tab:refcoco}. As a generalist model capable of handling complex instructions, our IVM achieves performance comparable to that of state-of-the-art (SOTA) specialist models.

\begin{table}[h]
    \centering
    \caption{result in REC}
    \begin{tabular}{lccc}
        \toprule
        Methods & RefCoCo & RefCoCo+ & RefCoCog \\
        \midrule
        \multicolumn{4}{c}{\textit{Specialist models}} \vspace{+4pt} \\
        G-DINO-L~\cite{liu2023grounding} & 90.56 & 82.75 & 86.13 \\
        \midrule
        \multicolumn{4}{c}{\textit{Generalist models}} \vspace{+4pt} \\
        LLaVA-7B~\cite{llava} & 76.29 &66.76 &70.4 \\
        % \midrule
        \rowcolor[gray]{.9} 
        IVM(Ours) & \textbf{90.1} & \textbf{83.3} & \textbf{82.9} \\
        \bottomrule
    \end{tabular}
    \label{tab:refcoco}
\end{table}

\subsection{Visualization Result}
\begin{figure}[h]
    \centering
    \includegraphics[width=1.0\textwidth]{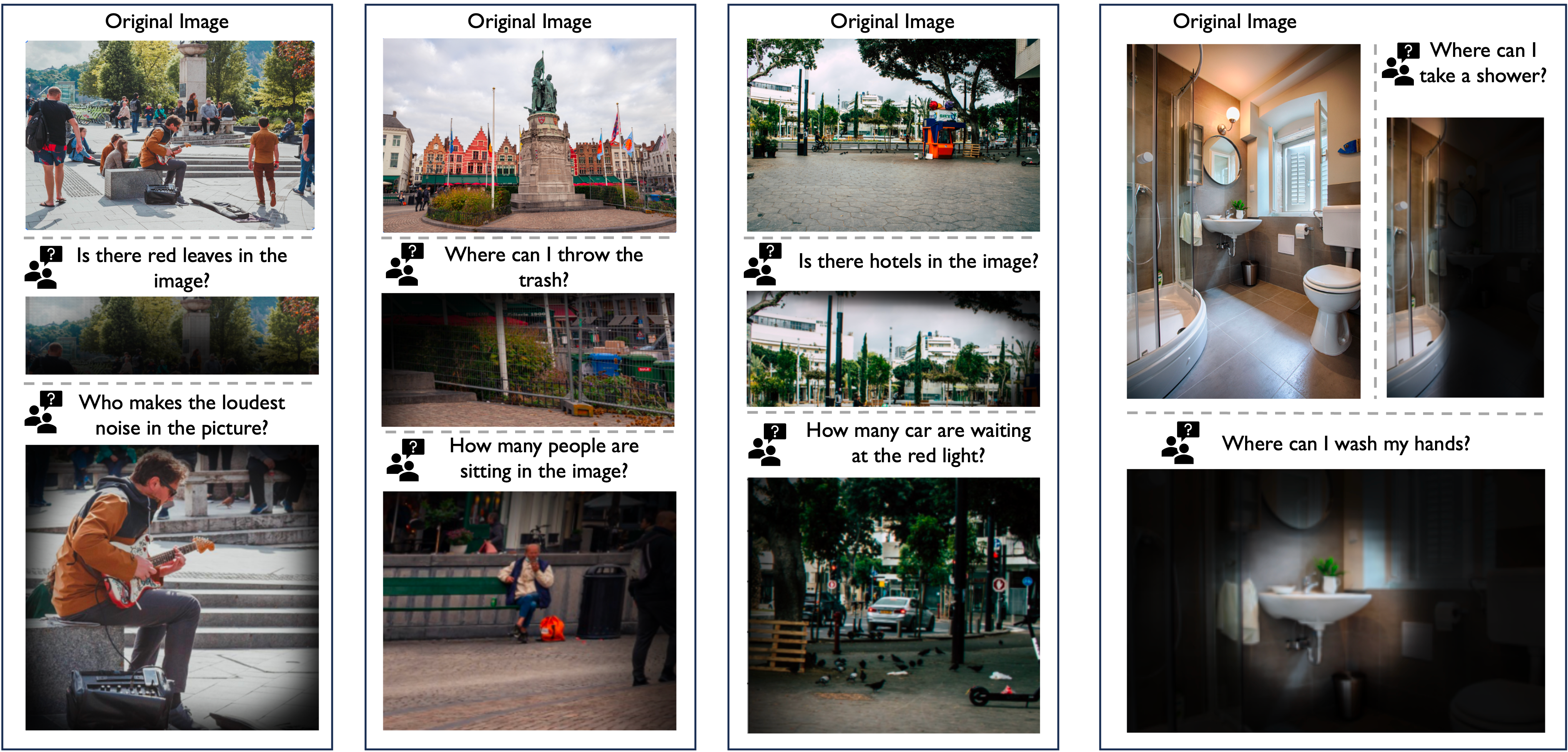}
    \caption{Visualization results of IVM generated masks.}
    \label{fig:goodcase}
\end{figure}

In this section, we provide more visualization result in VQA-type data as shown in Figure~\ref{fig:goodcase}. 

\begin{figure}[t]
    \centering
    \includegraphics[width=0.99\textwidth]{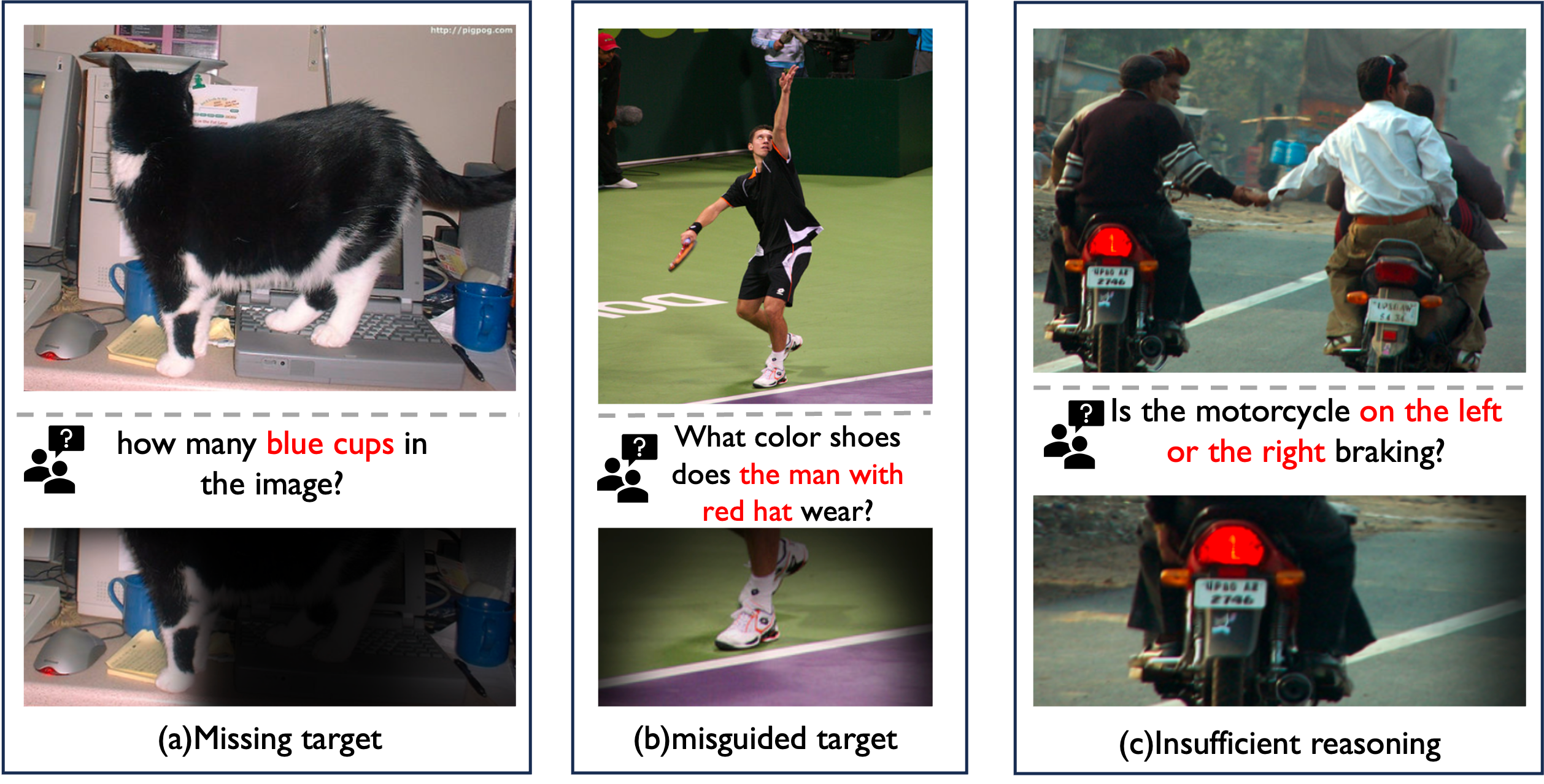}
    \caption{Some failure cases.}
    \label{fig:failure}
\end{figure}

\textbf{Failure Case}. Although we observe numerous successful instances, our IVM still faces significant challenges, as illustrated in Figure~\ref{fig:failure}. We summarize these challenges into three categories: missing target, misguided target, and insufficient reasoning.

(a) \textbf{Missing Target}: Challenges arise when target objects are relatively small and scattered around many separate image corners. In this case, accurately detecting all of the targeted objects is quite difficult. Even specialized open vocabulary detection models struggle with this task. For example, the cup on the right in the image is masked by the IVM mistakenly. However, we still observe that the IVM-generated heatmap for the right cup is partially activated, meaning that IVM have partially focus this regions. We believe by providing more training data, IVM can handle this better.

(b) \textbf{Misguided Target}: Accurately Localizing tiny target objects is a recognized challenge~\cite{wu2023vstar}, especially when similar but more obvious objects are present.
% Although IVM has shown impressive performance in such scenario, as demonstrated in the visualization in Figure~\ref{fig:goodcase}, errors still occur.
For instance, IVM incorrectly focuses on the more centrally located shoes of another man, instead of the shoes of the man wearing the \textbf{red hat} at the edge of the picture. However, this instruction is pretty challenging that at first glance, even a human might struggle to spot the man with the red hat in the left corner. We will leave challenge scenarios like this for future research.

(c) \textbf{Insufficient reasoning}: The objective of the IVM task is to assist LMMs in extracting visual features more effectively to better follow instructions. Thus, the demands on the model's reasoning capabilities extend far beyond mere object localization. Although IVM demonstrates strong performance, it sometimes overlooks additional image content necessary for accurately following instructions after correctly locating the target object. For instance, while IVM successfully identified the braking motorcycle, it failed to recognize that answering the question requires knowledge of the positions of both motorcycles simultaneously. We attribute this issue to biases in the training data. By incorporating more complex instructions and diversified labels, we anticipate that our model will achieve improved performance

\subsection{Robotics Result}
\begin{figure}[h]
    \centering
    \includegraphics[width=0.92\textwidth]{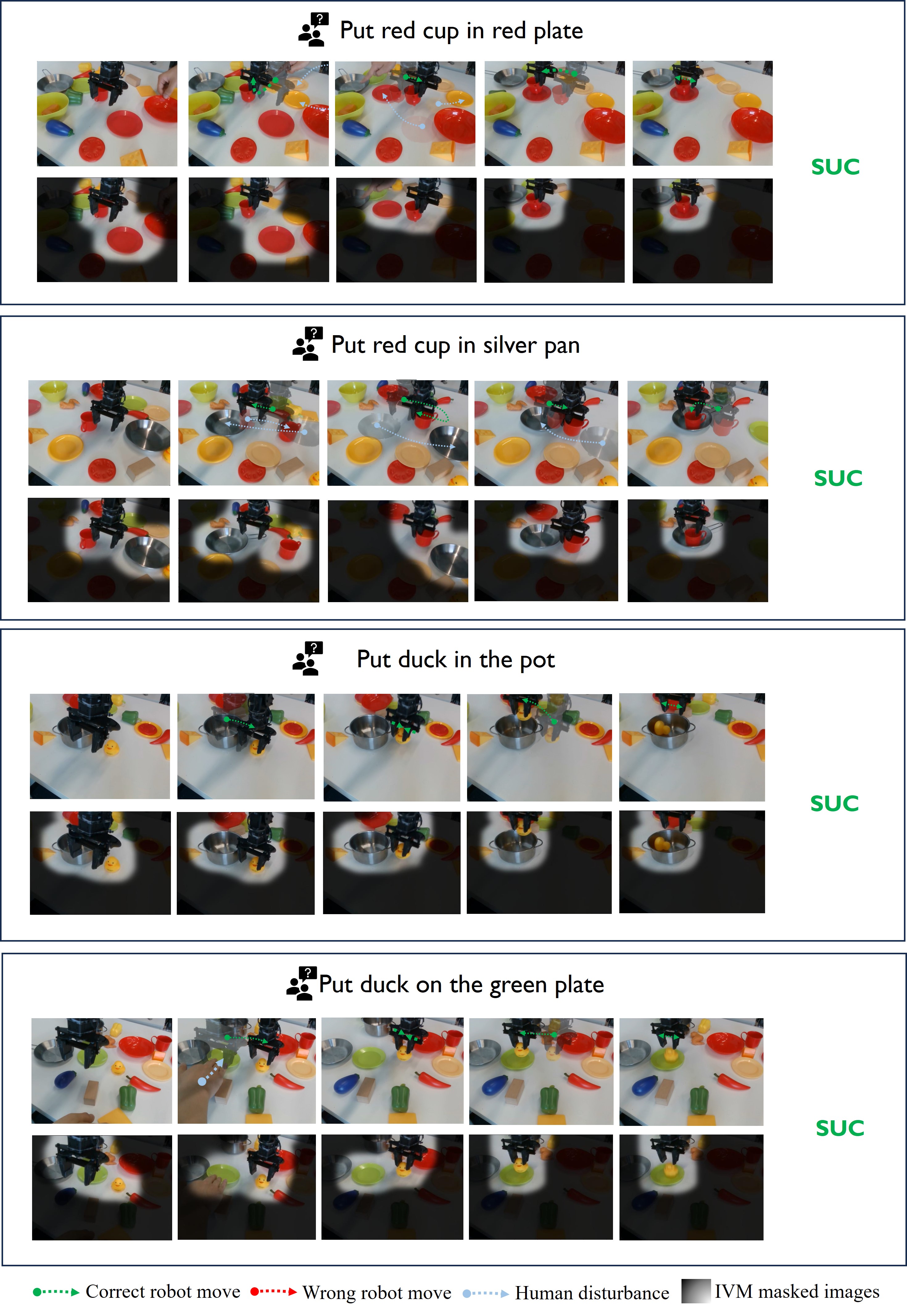}
    \caption{Real robot LCBC results with IVM assistance.}
    \label{fig:robot_exp_appen}
\end{figure}

Here, we provide more evaluation rollouts of the IVM-assisted LCBC agents under strong distractions. Figure~\ref{fig:robot_exp_appen} clearly demonstrates that even under strong distractions like the background are full of distracting objects with similar colors or shapes to the targeted objects, and strong human disturbances that adversarially attack the robots, the IVM-assisted LCBC agents can still complete the tasks pretty well, enjoying high-level of generalization and robustness thanks to the superior visual grounding ability injected by IVM. More videos can be found in the supplementary materials.

\newpage
\section*{NeurIPS Paper Checklist}

\begin{enumerate}

\item {\bf Claims}
    \item[] Question: Do the main claims made in the abstract and introduction accurately reflect the paper's contributions and scope?
    \item[] Answer: \answerYes{} % Replace by \answerYes{}, \answerNo{}, or \answerNA{}.
    \item[] Justification: Please see Abstract and Introduction for details.
    \item[] Guidelines:
    \begin{itemize}
        \item The answer NA means that the abstract and introduction do not include the claims made in the paper.
        \item The abstract and/or introduction should clearly state the claims made, including the contributions made in the paper and important assumptions and limitations. A No or NA answer to this question will not be perceived well by the reviewers. 
        \item The claims made should match theoretical and experimental results, and reflect how much the results can be expected to generalize to other settings. 
        \item It is fine to include aspirational goals as motivation as long as it is clear that these goals are not attained by the paper. 
    \end{itemize}

\item {\bf Limitations}
    \item[] Question: Does the paper discuss the limitations of the work performed by the authors?
    \item[] Answer: \answerYes{} % Replace by \answerYes{}, \answerNo{}, or \answerNA{}.
    \item[] Justification: Please see Conclusion and Appendix~\ref{sec:limitation} for details.
    \item[] Guidelines:
    \begin{itemize}
        \item The answer NA means that the paper has no limitation while the answer No means that the paper has limitations, but those are not discussed in the paper. 
        \item The authors are encouraged to create a separate "Limitations" section in their paper.
        \item The paper should point out any strong assumptions and how robust the results are to violations of these assumptions (e.g., independence assumptions, noiseless settings, model well-specification, asymptotic approximations only holding locally). The authors should reflect on how these assumptions might be violated in practice and what the implications would be.
        \item The authors should reflect on the scope of the claims made, e.g., if the approach was only tested on a few datasets or with a few runs. In general, empirical results often depend on implicit assumptions, which should be articulated.
        \item The authors should reflect on the factors that influence the performance of the approach. For example, a facial recognition algorithm may perform poorly when image resolution is low or images are taken in low lighting. Or a speech-to-text system might not be used reliably to provide closed captions for online lectures because it fails to handle technical jargon.
        \item The authors should discuss the computational efficiency of the proposed algorithms and how they scale with dataset size.
        \item If applicable, the authors should discuss possible limitations of their approach to address problems of privacy and fairness.
        \item While the authors might fear that complete honesty about limitations might be used by reviewers as grounds for rejection, a worse outcome might be that reviewers discover limitations that aren't acknowledged in the paper. The authors should use their best judgment and recognize that individual actions in favor of transparency play an important role in developing norms that preserve the integrity of the community. Reviewers will be specifically instructed to not penalize honesty concerning limitations.
    \end{itemize}

\item {\bf Theory Assumptions and Proofs}
    \item[] Question: For each theoretical result, does the paper provide the full set of assumptions and a complete (and correct) proof?
    \item[] Answer: \answerNA{} % Replace by \answerYes{}, \answerNo{}, or \answerNA{}.
    \item[] Justification: This paper has no theory.
    \item[] Guidelines:
    \begin{itemize}
        \item The answer NA means that the paper does not include theoretical results. 
        \item All the theorems, formulas, and proofs in the paper should be numbered and cross-referenced.
        \item All assumptions should be clearly stated or referenced in the statement of any theorems.
        \item The proofs can either appear in the main paper or the supplemental material, but if they appear in the supplemental material, the authors are encouraged to provide a short proof sketch to provide intuition. 
        \item Inversely, any informal proof provided in the core of the paper should be complemented by formal proofs provided in appendix or supplemental material.
        \item Theorems and Lemmas that the proof relies upon should be properly referenced. 
    \end{itemize}

    \item {\bf Experimental Result Reproducibility}
    \item[] Question: Does the paper fully disclose all the information needed to reproduce the main experimental results of the paper to the extent that it affects the main claims and/or conclusions of the paper (regardless of whether the code and data are provided or not)?
    \item[] Answer: \answerYes{} % Replace by \answerYes{}, \answerNo{}, or \answerNA{}.
    \item[] Justification: Please see Appendix~\ref{sec:train_eval_appenx} for details.
    \item[] Guidelines:
    \begin{itemize}
        \item The answer NA means that the paper does not include experiments.
        \item If the paper includes experiments, a No answer to this question will not be perceived well by the reviewers: Making the paper reproducible is important, regardless of whether the code and data are provided or not.
        \item If the contribution is a dataset and/or model, the authors should describe the steps taken to make their results reproducible or verifiable. 
        \item Depending on the contribution, reproducibility can be accomplished in various ways. For example, if the contribution is a novel architecture, describing the architecture fully might suffice, or if the contribution is a specific model and empirical evaluation, it may be necessary to either make it possible for others to replicate the model with the same dataset, or provide access to the model. In general. releasing code and data is often one good way to accomplish this, but reproducibility can also be provided via detailed instructions for how to replicate the results, access to a hosted model (e.g., in the case of a large language model), releasing of a model checkpoint, or other means that are appropriate to the research performed.
        \item While NeurIPS does not require releasing code, the conference does require all submissions to provide some reasonable avenue for reproducibility, which may depend on the nature of the contribution. For example
        \begin{enumerate}
            \item If the contribution is primarily a new algorithm, the paper should make it clear how to reproduce that algorithm.
            \item If the contribution is primarily a new model architecture, the paper should describe the architecture clearly and fully.
            \item If the contribution is a new model (e.g., a large language model), then there should either be a way to access this model for reproducing the results or a way to reproduce the model (e.g., with an open-source dataset or instructions for how to construct the dataset).
            \item We recognize that reproducibility may be tricky in some cases, in which case authors are welcome to describe the particular way they provide for reproducibility. In the case of closed-source models, it may be that access to the model is limited in some way (e.g., to registered users), but it should be possible for other researchers to have some path to reproducing or verifying the results.
        \end{enumerate}
    \end{itemize}

\item {\bf Open access to data and code}
    \item[] Question: Does the paper provide open access to the data and code, with sufficient instructions to faithfully reproduce the main experimental results, as described in supplemental material?
    \item[] Answer: \answerYes{} % Replace by \answerYes{}, \answerNo{}, or \answerNA{}.
    \item[] Justification: Code, model and data are available at \href{https://github.com/2toinf/IVM}{\texttt{https://github.com/2toinf/IVM}}.
    \item[] Guidelines:
    \begin{itemize}
        \item The answer NA means that paper does not include experiments requiring code.
        \item Please see the NeurIPS code and data submission guidelines (\url{https://nips.cc/public/guides/CodeSubmissionPolicy}) for more details.
        \item While we encourage the release of code and data, we understand that this might not be possible, so “No” is an acceptable answer. Papers cannot be rejected simply for not including code, unless this is central to the contribution (e.g., for a new open-source benchmark).
        \item The instructions should contain the exact command and environment needed to run to reproduce the results. See the NeurIPS code and data submission guidelines (\url{https://nips.cc/public/guides/CodeSubmissionPolicy}) for more details.
        \item The authors should provide instructions on data access and preparation, including how to access the raw data, preprocessed data, intermediate data, and generated data, etc.
        \item The authors should provide scripts to reproduce all experimental results for the new proposed method and baselines. If only a subset of experiments are reproducible, they should state which ones are omitted from the script and why.
        \item At submission time, to preserve anonymity, the authors should release anonymized versions (if applicable).
        \item Providing as much information as possible in supplemental material (appended to the paper) is recommended, but including URLs to data and code is permitted.
    \end{itemize}

\item {\bf Experimental Setting/Details}
    \item[] Question: Does the paper specify all the training and test details (e.g., data splits, hyperparameters, how they were chosen, type of optimizer, etc.) necessary to understand the results?
    \item[] Answer: \answerYes{} % Replace by \answerYes{}, \answerNo{}, or \answerNA{}.
    \item[] Justification: Please see Appendix~\ref{sec:train_eval_appenx} for details.
    \item[] Guidelines:
    \begin{itemize}
        \item The answer NA means that the paper does not include experiments.
        \item The experimental setting should be presented in the core of the paper to a level of detail that is necessary to appreciate the results and make sense of them.
        \item The full details can be provided either with the code, in appendix, or as supplemental material.
    \end{itemize}

\item {\bf Experiment Statistical Significance}
    \item[] Question: Does the paper report error bars suitably and correctly defined or other appropriate information about the statistical significance of the experiments?
    \item[] Answer: \answerNo{} % Replace by \answerYes{}, \answerNo{}, or \answerNA{}.
    \item[] Justification: Error bars are not reported because it would be too computationally expensive.
    \item[] Guidelines:
    \begin{itemize}
        \item The answer NA means that the paper does not include experiments.
        \item The authors should answer "Yes" if the results are accompanied by error bars, confidence intervals, or statistical significance tests, at least for the experiments that support the main claims of the paper.
        \item The factors of variability that the error bars are capturing should be clearly stated (for example, train/test split, initialization, random drawing of some parameter, or overall run with given experimental conditions).
        \item The method for calculating the error bars should be explained (closed form formula, call to a library function, bootstrap, etc.)
        \item The assumptions made should be given (e.g., Normally distributed errors).
        \item It should be clear whether the error bar is the standard deviation or the standard error of the mean.
        \item It is OK to report 1-sigma error bars, but one should state it. The authors should preferably report a 2-sigma error bar than state that they have a 96\% CI, if the hypothesis of Normality of errors is not verified.
        \item For asymmetric distributions, the authors should be careful not to show in tables or figures symmetric error bars that would yield results that are out of range (e.g. negative error rates).
        \item If error bars are reported in tables or plots, The authors should explain in the text how they were calculated and reference the corresponding figures or tables in the text.
    \end{itemize}

\item {\bf Experiments Compute Resources}
    \item[] Question: For each experiment, does the paper provide sufficient information on the computer resources (type of compute workers, memory, time of execution) needed to reproduce the experiments?
    \item[] Answer: \answerYes{} % Replace by \answerYes{}, \answerNo{}, or \answerNA{}.
    \item[] Justification: Please see Appendix~\ref{sec:train_eval_appenx} for details.
    \item[] Guidelines:
    \begin{itemize}
        \item The answer NA means that the paper does not include experiments.
        \item The paper should indicate the type of compute workers CPU or GPU, internal cluster, or cloud provider, including relevant memory and storage.
        \item The paper should provide the amount of compute required for each of the individual experimental runs as well as estimate the total compute. 
        \item The paper should disclose whether the full research project required more compute than the experiments reported in the paper (e.g., preliminary or failed experiments that didn't make it into the paper). 
    \end{itemize}
    
\item {\bf Code Of Ethics}
    \item[] Question: Does the research conducted in the paper conform, in every respect, with the NeurIPS Code of Ethics \url{https://neurips.cc/public/EthicsGuidelines}?
    \item[] Answer: \answerYes{} % Replace by \answerYes{}, \answerNo{}, or \answerNA{}.
    \item[] Justification: N/A
    \item[] Guidelines:
    \begin{itemize}
        \item The answer NA means that the authors have not reviewed the NeurIPS Code of Ethics.
        \item If the authors answer No, they should explain the special circumstances that require a deviation from the Code of Ethics.
        \item The authors should make sure to preserve anonymity (e.g., if there is a special consideration due to laws or regulations in their jurisdiction).
    \end{itemize}

\item {\bf Broader Impacts}
    \item[] Question: Does the paper discuss both potential positive societal impacts and negative societal impacts of the work performed?
    \item[] Answer: \answerYes{} % Replace by \answerYes{}, \answerNo{}, or \answerNA{}.
    \item[] Justification: Please see Appendix~\ref{sec:broad_impact} for details.
    \item[] Guidelines:
    \begin{itemize}
        \item The answer NA means that there is no societal impact of the work performed.
        \item If the authors answer NA or No, they should explain why their work has no societal impact or why the paper does not address societal impact.
        \item Examples of negative societal impacts include potential malicious or unintended uses (e.g., disinformation, generating fake profiles, surveillance), fairness considerations (e.g., deployment of technologies that could make decisions that unfairly impact specific groups), privacy considerations, and security considerations.
        \item The conference expects that many papers will be foundational research and not tied to particular applications, let alone deployments. However, if there is a direct path to any negative applications, the authors should point it out. For example, it is legitimate to point out that an improvement in the quality of generative models could be used to generate deepfakes for disinformation. On the other hand, it is not needed to point out that a generic algorithm for optimizing neural networks could enable people to train models that generate Deepfakes faster.
        \item The authors should consider possible harms that could arise when the technology is being used as intended and functioning correctly, harms that could arise when the technology is being used as intended but gives incorrect results, and harms following from (intentional or unintentional) misuse of the technology.
        \item If there are negative societal impacts, the authors could also discuss possible mitigation strategies (e.g., gated release of models, providing defenses in addition to attacks, mechanisms for monitoring misuse, mechanisms to monitor how a system learns from feedback over time, improving the efficiency and accessibility of ML).
    \end{itemize}
    
\item {\bf Safeguards}
    \item[] Question: Does the paper describe safeguards that have been put in place for responsible release of data or models that have a high risk for misuse (e.g., pretrained language models, image generators, or scraped datasets)?
    \item[] Answer: \answerYes{} % Replace by \answerYes{}, \answerNo{}, or \answerNA{}.
    \item[] Justification: All data are collected from open-sourced and peer-reviewed dataset. The models used for annotations are also open-sourced  and peer-reviewed.
    \item[] Guidelines:
    \begin{itemize}
        \item The answer NA means that the paper poses no such risks.
        \item Released models that have a high risk for misuse or dual-use should be released with necessary safeguards to allow for controlled use of the model, for example by requiring that users adhere to usage guidelines or restrictions to access the model or implementing safety filters. 
        \item Datasets that have been scraped from the Internet could pose safety risks. The authors should describe how they avoided releasing unsafe images.
        \item We recognize that providing effective safeguards is challenging, and many papers do not require this, but we encourage authors to take this into account and make a best faith effort.
    \end{itemize}

\item {\bf Licenses for existing assets}
    \item[] Question: Are the creators or original owners of assets (e.g., code, data, models), used in the paper, properly credited and are the license and terms of use explicitly mentioned and properly respected?
    \item[] Answer: \answerYes{} % Replace by \answerYes{}, \answerNo{}, or \answerNA{}.
    \item[] Justification: N/A
    \item[] Guidelines:
    \begin{itemize}
        \item The answer NA means that the paper does not use existing assets.
        \item The authors should cite the original paper that produced the code package or dataset.
        \item The authors should state which version of the asset is used and, if possible, include a URL.
        \item The name of the license (e.g., CC-BY 4.0) should be included for each asset.
        \item For scraped data from a particular source (e.g., website), the copyright and terms of service of that source should be provided.
        \item If assets are released, the license, copyright information, and terms of use in the package should be provided. For popular datasets, \url{paperswithcode.com/datasets} has curated licenses for some datasets. Their licensing guide can help determine the license of a dataset.
        \item For existing datasets that are re-packaged, both the original license and the license of the derived asset (if it has changed) should be provided.
        \item If this information is not available online, the authors are encouraged to reach out to the asset's creators.
    \end{itemize}

\item {\bf New Assets}
    \item[] Question: Are new assets introduced in the paper well documented and is the documentation provided alongside the assets?
    \item[] Answer: \answerYes{} % Replace by \answerYes{}, \answerNo{}, or \answerNA{}.
    \item[] Justification: We will release our IVM-Mix-1M dataset and detailed documentations after the acceptance.
    \item[] Guidelines:
    \begin{itemize}
        \item The answer NA means that the paper does not release new assets.
        \item Researchers should communicate the details of the dataset/code/model as part of their submissions via structured templates. This includes details about training, license, limitations, etc. 
        \item The paper should discuss whether and how consent was obtained from people whose asset is used.
        \item At submission time, remember to anonymize your assets (if applicable). You can either create an anonymized URL or include an anonymized zip file.
    \end{itemize}

\item {\bf Crowdsourcing and Research with Human Subjects}
    \item[] Question: For crowdsourcing experiments and research with human subjects, does the paper include the full text of instructions given to participants and screenshots, if applicable, as well as details about compensation (if any)? 
    \item[] Answer: \answerYes{} % Replace by \answerYes{}, \answerNo{}, or \answerNA{}.
    \item[] Justification: All crowdsourcing labels in the IVM-Mix-1M dataset are annotated by the authors.
    \item[] Guidelines:
    \begin{itemize}
        \item The answer NA means that the paper does not involve crowdsourcing nor research with human subjects.
        \item Including this information in the supplemental material is fine, but if the main contribution of the paper involves human subjects, then as much detail as possible should be included in the main paper. 
        \item According to the NeurIPS Code of Ethics, workers involved in data collection, curation, or other labor should be paid at least the minimum wage in the country of the data collector. 
    \end{itemize}

\item {\bf Institutional Review Board (IRB) Approvals or Equivalent for Research with Human Subjects}
    \item[] Question: Does the paper describe potential risks incurred by study participants, whether such risks were disclosed to the subjects, and whether Institutional Review Board (IRB) approvals (or an equivalent approval/review based on the requirements of your country or institution) were obtained?
    \item[] Answer: \answerNA{} % Replace by \answerYes{}, \answerNo{}, or \answerNA{}.
    \item[] Justification: This paper has no human subjects.
    \item[] Guidelines:
    \begin{itemize}
        \item The answer NA means that the paper does not involve crowdsourcing nor research with human subjects.
        \item Depending on the country in which research is conducted, IRB approval (or equivalent) may be required for any human subjects research. If you obtained IRB approval, you should clearly state this in the paper. 
        \item We recognize that the procedures for this may vary significantly between institutions and locations, and we expect authors to adhere to the NeurIPS Code of Ethics and the guidelines for their institution. 
        \item For initial submissions, do not include any information that would break anonymity (if applicable), such as the institution conducting the review.
    \end{itemize}

\end{enumerate}

% \section*{References}

%%%%%%%%%%%%%%%%%%%%%%%%%%%%%%%%%%%%%%%%%%%%%%%%%%%%%%%%%%%%

\end{document}